\documentclass[10pt,twocolumn,letterpaper]{article}

\usepackage{cvpr} 

\usepackage{xcolor}

\newcommand{\ModelNameShort}{SCD}
\newcommand{\Encoder}{\mathcal{E}_{\phi}}
\newcommand{\Decoder}{\mathcal{D}_{\theta}}

\usepackage{microtype}

\makeatletter
\renewcommand\section{\@startsection{section}{1}{\z@}%
  {-2.5ex plus -1ex minus -.2ex}%
  {1.8ex plus .2ex}%
  {\normalfont\large\bfseries}}
\renewcommand\subsection{\@startsection{subsection}{2}{\z@}%
  {-2ex plus -1ex minus -.2ex}%
  {1.2ex plus .2ex}%
  {\normalfont\normalsize\bfseries}}
\makeatother

\definecolor{cvprblue}{rgb}{0.21,0.49,0.74}
\usepackage[pagebackref,breaklinks,colorlinks,allcolors=cvprblue]{hyperref}
\usepackage{multirow}
\usepackage{tikz}
\usepackage[most]{tcolorbox}
\usepackage{booktabs}      
\usepackage[table]{xcolor} 
\definecolor{catgray}{RGB}{244,244,244} 
\tcbuselibrary{skins, breakable, theorems}
\usepackage{algorithm}            
\usepackage[noend]{algpseudocode} 
\usepackage{amsmath,amssymb,mathtools}
\algrenewcommand\algorithmicrequire{\textbf{Require:}}
\algrenewcommand\algorithmicensure{\textbf{Output:}}
\algrenewcommand\algorithmiccomment[1]{\hfill$\triangleright$~#1}
\definecolor{scholarblue}{RGB}{33,93,156}
\newtcolorbox{styledquote}[1][]{%
  enhanced,
  breakable,
  colback=scholarblue!8,     
  colframe=white,            
  boxrule=0pt,
  sharp corners,
  fontupper=\itshape,        
  left=6pt, right=6pt, top=3pt, bottom=3pt,
  width=\columnwidth,        
  before skip=8pt, after skip=8pt,
  #1
}

\newcommand{\mymodel}{Separable Causal Diffusion\ }
\newcommand{\mymodelab}{SCD}

\def\paperID{*****} 
\def\confName{CVPR}
\def\confYear{2026}

\title{
Causality in Video Diffusers is Separable from Denoising

}

\author{Xingjian Bai$^{1,2}$\thanks{Work done while Xingjian was interning at Adobe.} \quad Guande He$^{3}$ \quad Zhengqi Li$^2$\\
Eli Shechtman$^2$ \quad Xun Huang$^3$ \quad Zongze Wu$^2$\\
$^1$Massachusetts Institute of Technology \quad $^2$Adobe Research \\
 $^3$Morpheus AI
}

\begin{document}
\maketitle

\begin{abstract}
Causality --- referring to temporal, uni-directional cause-effect relationships between components --- underlies many complex generative processes, including videos, language, and robot trajectories.
Current causal diffusion models entangle temporal reasoning with iterative denoising, applying causal attention across all layers, at every denoising step, and over the entire context.
In this paper, we show that the causal reasoning in these models is separable from the multi-step denoising process.
Through systematic probing of autoregressive video diffusers, we uncover two key regularities:
(1) early layers produce highly similar features across denoising steps, indicating redundant computation along the diffusion trajectory; and
(2) deeper layers exhibit sparse cross-frame attention and primarily perform intra-frame rendering.
Motivated by these findings, we introduce \mymodel (\mymodelab), a new architecture that explicitly decouples once-per-frame temporal reasoning, via a causal transformer encoder, from multi-step frame‑wise rendering, via a lightweight diffusion decoder.
Extensive experiments on both pretraining and post-training tasks across synthetic and real benchmarks show that \mymodelab\  significantly improves throughput and per-frame latency while matching or surpassing the generation quality of strong causal diffusion baselines.

\end{abstract}
    
\section{Introduction}

Modeling causality\footnote{In this paper we use \emph{causality} narrowly to mean the temporal arrow-of-time—\emph{the past determines the future, not vice versa.}} is a core problem in diffusion generation modeling. 
Starting from fitting image distributions \cite{rombach2022high, podell2023sdxl, ramesh2022dalle2, openai2023dalle3, balaji2022ediffi, saharia2022imagen, chen2023pixart, esser2024sd3}, 
diffusion models \cite{ho2020denoising, lipman2022flowmatching} have achieved great success across modalities such as videos~\cite{ho2022vdm, ho2022imagen, yang2024cogvideox, BarTal2024LumiereAS, guo2023animatediff, wang2023modelscopet2v, wan2025}, audio~\cite{kong2021diffwave, liu2023audioldm}, and language~\cite{HeSunWangEtAl2022_DiffusionBERT, lovelace2022ld4lg, NieZhuYouEtAl2025_LLaDA}. 
In its basic form, diffusion models denoise all tokens simultaneously, generating the entire output all at once. This is still the design of many state-of-the-art video diffusion models. However, this formulation overlooks the temporal evolution inherent in sequential data—allowing the future information to influence the past, and preventing crucial applications such as long-term, real-time video streaming~\cite{Yan2021VideoGPT, kondratyuk2024videopoet, yin2025causvid, huang2025selfforcing}. 
To incorporate temporal causal dependencies and enable autoregressive video generation, researchers have attempted to replace bidirectional full attention inside the denoiser with causal attention~\cite{yin2025causvid, gu2025far, huang2025selfforcing}, as commonly used in the LLM community. This mechanism applies bidirectional attention within a frame (or chunk of frames) and causal attention across frames (or chunks). When combined with the diffusion process, every token within a frame must pass through the entire network iteratively, computing both intra-frame and cross-frame attention at every layer and every denoising step.

\begin{figure}[t] 
    \centering
    \includegraphics[width=\linewidth]{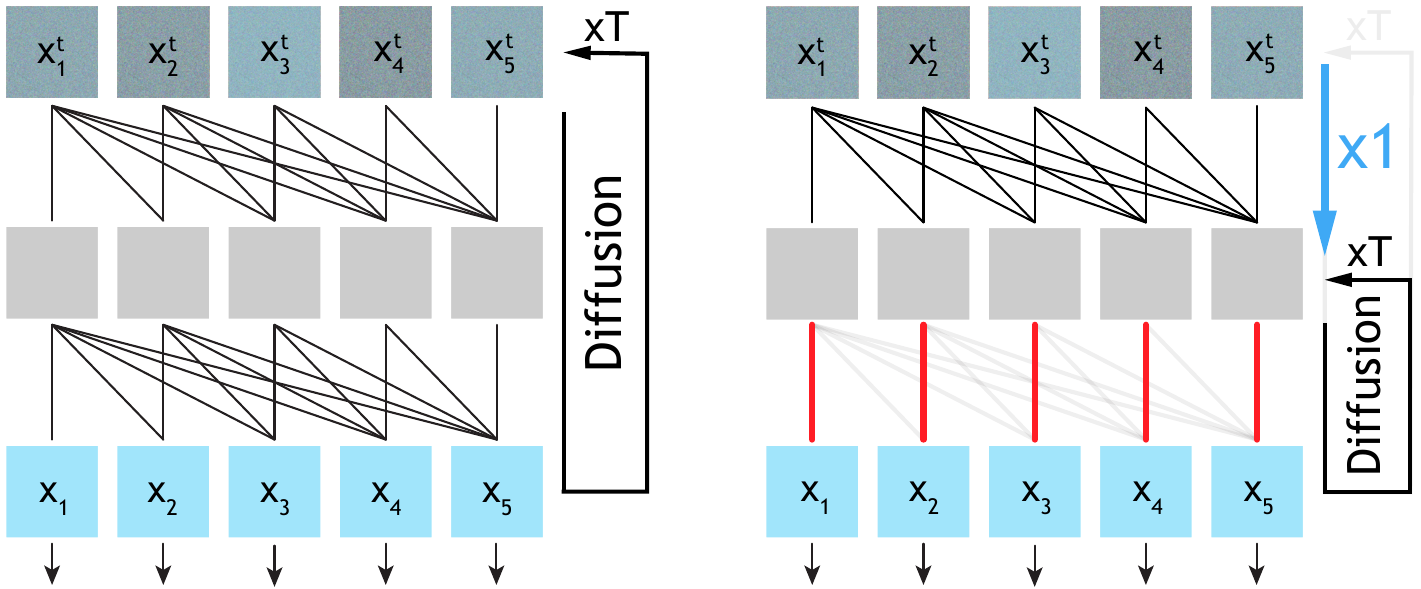}
    \caption{
    \textbf{Causality in autoregressive video diffusion models is separable from the denoising process.} 
    The prevailing design of causal diffusion models for visual generation performs causal attention densely across \emph{all layers and all denoising steps} (left).
    However, we uncover two important observations (right): 1) early denoiser layers share highly repetitive computation across denoising steps (blue); 2) deep layers primarily attend to intra-frame tokens, with sparse cross-frame connections (red).}
    \label{fig:teaser}
    \vspace{-1em}
\end{figure}

While causal attention is essential for modeling temporal evolution, directly transplanting it from LLMs overlooks a key difference: diffusion models typically perform multi-step refinement for each frame, rather than generating in a single pass. The current design of causal diffusion tightly entangles temporal reasoning with iterative denoising, with each layer at every step repeatedly performing causal reasoning. This raises a fundamental question: Is multi-step refinement truly required for temporal reasoning?

To answer this question, we conduct detailed probing analysis and finetuning experiments on autoregressive (AR) video diffusion models. We consistently observe that temporal reasoning in AR models is separable from the denoising process (Fig.~\ref{fig:teaser}).
In particular, we find that causal reasoning in early layers is highly redundant across denoising timesteps, as indicated by the high similarity in middle-layer output features across denoising steps. We also observe that temporal computation in deeper layers is far less frequent: careful attention visualizations reveal that deeper layers predominantly perform intra-frame attention while rarely attending across frames.




Motivated by the sparsity and redundancy we uncover, we introduce \mymodel (\mymodelab), a novel decoupled causal architecture in which a temporal causal-reasoning module operates once per frame, while a lightweight frame‑wise diffusion renderer handles visual refinement. Concretely, a causal transformer reads the historical clean frame tokens through KV cache and produces a latent that summarizes the entities, layout, and expected motion from its context. This context latent is then reused across all denoising steps for that frame.
A diffusion module receives both the current noisy frame tokens and the context latent, and performs a frame‑wise iterative denoising process without any cross-frame computation.
Taken together, our design mirrors next-token prediction in LLMs (except that we perform next-\textit{frame} prediction here followed by continuous rendering), reallocating compute from repeated cross-frame operations to per-frame refinement, thereby reducing latency and memory while preserving generation quality.



We conduct extensive experiments at both pretraining and post-training stages for causal video diffusion models across synthetic and real datasets. We show that \mymodelab{} trained from scratch matches or surpasses causal diffusion baselines in generation quality while achieving 2–3× lower latency.
Furthermore, to demonstrate scalability, we finetune \mymodelab~from a pretrained bidirectional teacher diffusion model, achieving strong video generation quality with substantially higher throughput compared with AR baselines.

\paragraph{Contributions.} In summary, we make the following contributions: 	1) Through careful probing and finetuning experiments, we observe that causal reasoning in existing causal video diffusion models is redundant across denoising steps and sparse across time. 2) We introduce a novel \mymodel (\mymodelab) architecture that fully leverages these observations. On both pretraining and post-training tasks, \mymodelab~demonstrates strong effectiveness across multiple datasets compared with baseline models.


\section{Related Work}

\label{sec:related_work}

\noindent\textbf{From Bidirectional to Autoregressive Video Diffusion.} 
Diffusion-based video generative models have achieved remarkable fidelity by employing spatio-temporal Transformers with bidirectional attention over entire video sequences. Recent methods~\cite{Gupta2023PhotorealisticVG, BarTal2024LumiereAS, Polyak2024MovieGA, Chen2023VideoCrafter1, deepmind_veo3_techreport, wan2025, Gao2025Seedance1E, chen2025skyreelsv2, deng2024nova} advance this paradigm through careful architectural design and large-scale training, achieving state-of-the-art visual quality. However, their non-causal design requires generating all frames simultaneously, resulting in high latency and preventing real-time streaming or interactive applications.

To enable online, low-latency generation, recent efforts have shifted toward autoregressive (AR) video generation, particularly using diffusion models with causal transformers. Instead of producing all frames at once, AR diffusion models generate videos in a causal manner, conditioning each frame only on past frames. This causal dependence not only aligns with the arrow of time but also enables efficient inference via KV caching, making it attractive for interactive settings. Pioneering AR approaches include models trained from scratch~\cite{gu2025far, chen2025skyreelsv2, sandai2025magi1, oshima2024ssmvdm, kondratyuk2024videopoet} and techniques that distill a causal generator from a pretrained video diffusion model~\cite{chen2024diffusionforcing, yang2025longlive, huang2025selfforcing, yin2025causvid, cui2025selfforcingpp}.

\noindent\textbf{AR-Diffusion Hybrid Models.} 
To leverage the strengths of both paradigms, a growing body of work combines an AR module with a diffusion module. In the image domain, several recent works~\cite{li2024mar, Fan2025UnifiedAV} have demonstrated that an AR transformer can operate on continuous tokens to generate a coarse layout, which is then refined by a diffusion module to produce high-fidelity images.
In the video domain, MarDini~\cite{liu2024mardini} and VideoMAR~\cite{yu2025videomar} both employ an AR module to produce a context representation of the video, which is subsequently used by a diffusion module to generate visual tokens.
Notably, VideoPoet~\cite{kondratyuk2024videopoet} also adopts a frame‑wise autoregressive strategy, but it uses a single-pass decoder operating on discrete tokens and lacks a diffusion module for refinement, leading to low-quality generation.
In parallel, another line of work aims to unify understanding and generation tasks through hybrid AR transformers paired with diffusion heads~\cite{Fan2025UnifiedAV, Tong2024MetaMorphMU, Zhou2024TransfusionPT, Mo2025XFusionIN, Shi2024LMFusionAP}.


\noindent\textbf{Separability and Sparsity in Video Models.}
Separability and sparsity have long been central themes in video modeling: because the space-time dimension is dense, naively porting image architectures becomes prohibitive, motivating early/late fusion and factorized designs that decouple spatial and temporal processing~\cite{karpathy2014largevideocnn, simonyan2014twostream, carreira2017i3d, tran2018r2plus1d, lin2019tsm, bertasius2021timesformer, arnab2021vivit, liu2022videoswin}. In video diffusion models, researchers have recently leveraged inherent 3D attention patterns from pretrained video models to accelerate generation~\cite{Xi2025SparseVA, Yang2025SparseVA, Zhang2025VSAFV, Zhang2025FastVG}. Our work can be viewed as a continuation of this discussion on separability and sparsity in video models, specifically in the context of temporally causal video diffusion.
Beyond video, separability in diffusion models has likewise been studied and exploited in other modalities, including images~\cite{wang2025ddt} and language~\cite{arriola2025encoderdecoderdiffusionlanguagemodels}.



\section{Preliminaries: Causal Diffusion Models}
\label{sec:3.1}


In this section, we review the causal diffusion paradigm, a variant of diffusion models that generates a step-indexed sequence in a causal manner. This is the predominant pipeline for frame-autoregressive video generation \cite{gu2025far, chen2024diffusionforcing, huang2025selfforcing, yin2025causvid}. We formalize the continuous-time objective and highlight why it embeds causal dependence throughout the entire diffusion trajectory. Finally, we briefly introduce Teacher Forcing and Diffusion Forcing~\cite{chen2024diffusionforcing} 
as training techniques for causal diffusion models.


A causal generator models the joint distribution of a sequence \(x_{1:N}=(x_1,\dots,x_N)\) by predicting each element from its past. With optional per-step controls \(a_{1:N}\)\footnote{Global controls (e.g., a text prompt) can be treated as constant per-step conditioning.}, the joint distribution factorizes as
\begin{equation}
\label{eq:ar_factorization}
p_\theta(x_{1:N} \mid a_{1:N}) \;=\; \prod_{i=1}^{N} p_\theta\!\big(x_i \mid C_i= (x_{<i},\, a_{\le i})\big),
\end{equation}
where each conditional probability is implemented by a diffusion renderer that attends to its context, 
\(C_i\).

We adopt a continuous notion of time, \(t\in[0,1]\) and define a forward diffusion path, connecting the data distribution with a standard Gaussian \(\mathcal{N}(0,I)\):
\begin{equation}
\label{eq:diffusion_forward}
x_i^t \;=\; (1-t)\,x_i \;+\; t\,\epsilon_i \text{, where } \epsilon_i\sim\mathcal{N}(0,I).
\end{equation}
A causal diffusion network takes the noisy samples as input, and \(v_\theta\) predicts its velocity on the diffusion path, conditioning on \((t, C_i)\) 
\[
\hat{v}_{i,\theta} \;=\; v_\theta(x_i^t,\; t,\; C_i)\,,
\]
while the ground-truth velocity is the time derivative of the diffusion path
\begin{equation}
\label{eq:cond_velocity}
u(x_i^t,\, t \mid x_i) \;=\; \frac{d}{dt}x_i^t \;=\; \epsilon_i - x_i.
\end{equation}

Training loss is defined on the gap between the predicted and ground-truth velocity under a time-weighted expectation:
\begingroup\small
\begin{equation}
\label{eq:loss}
L(\theta)
=\mathbb{E}_{\,x,\,i,\,t,\,\epsilon}\!\Big[\,w(t)\,
\big\|\,u(x_i^t,\,t\mid x_i)-v_\theta(x_i^t,\,t,\,C_i)\,\big\|^2\Big],
\end{equation}
\endgroup
where \(w(t)\) is a standard time weighting.  
 Crucially, because \(v_\theta(\cdot,t,C_i)\) is conditioned on \(C_i\) \emph{for every} \(t\in[0,1]\) and \(L(\theta)\) integrates over \(t\), the model must repeatedly consult the context along the whole reverse trajectory: causal reasoning is therefore entangled with the entire diffusion path (and, in common implementations of the denoiser, propagated across all denoiser layers at each step).

\paragraph{Teacher Forcing and Diffusion Forcing.}

\textit{Teacher Forcing (TF)} trains next-frame prediction with \emph{clean} history: at each step the denoiser predicts the current frame while attending to ground-truth context frames. 
This provides a standard causal diffusion training recipe but induces a train–test mismatch: at inference, the model conditions on its own imperfect past outputs, resulting in severe error accumulation during roll-out~\cite{chen2024diffusionforcing, song2025historyguidedvideodiffusion}.
\textit{Diffusion Forcing (DF)}~\cite{chen2024diffusionforcing} addresses this by \emph{noising the context} during training: each context frame is independently perturbed to a sampled noise level, and the denoiser predicts the current frame while attending to these partially noised contexts. This simple augmentation better matches inference-time erroneous conditions. However, diffusion forcing conditions on noisy  ground truth inputs during training but relies on clean past rollout at inference, leading to another form of mismatch between training and test conditions.

\begin{figure*}[t] 
    \centering
    \includegraphics[width=\textwidth]{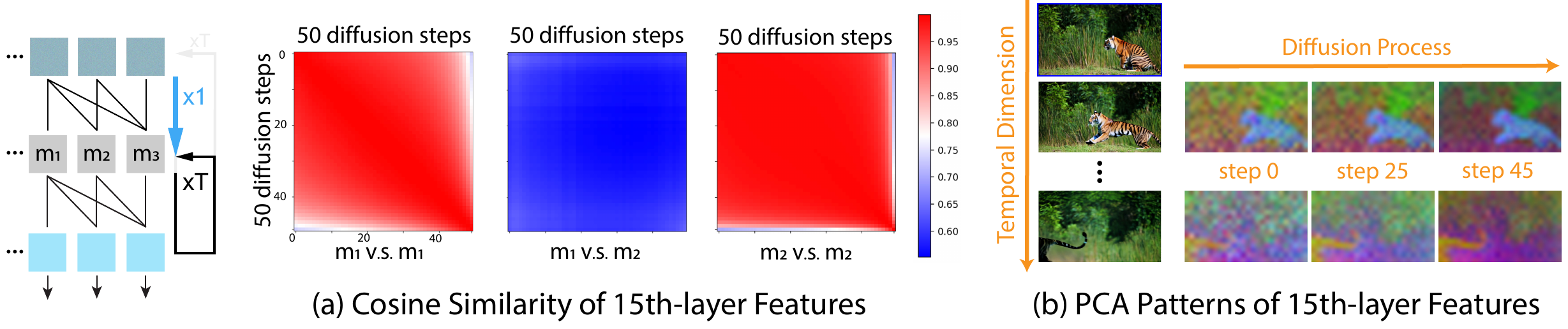}
    \caption{
    \textbf{Strong middle-block feature consistency across denoising steps.}
    (a) When denoising the same frame over 50 steps, the middle-block (15th block out of 30) features exhibit consistently high cosine similarity (above 0.95), suggesting that the features generated in the middle block are mostly shared across different diffusion steps.
    (b) PCA analysis further confirms that the middle-block features at the first and later diffusion steps are highly aligned, indicating that structures are effectively established even in the first step.
    }
    \label{fig:obs2_combined_motivational}
    \vspace{-1.5em}
\end{figure*}

\section{Uncovering Causal Separability}
\label{sec:analyses}

In this section, we study where the main \emph{causal reasoning} actually occurs inside AR video diffusers. As a testbed, we adopt \textit{WAN‑2.1 T2V‑1.3B} ~\cite{wan2025}, one of the most capable open‑source text‑to‑video models, and convert it to a frame‑wise AR generator via teacher forcing \cite{yin2025causvid,wan2025}.
To ensure that our findings do not hinge on this particular choice, we repeat similar observations on autoregressive video models trained from scratch and on other conditioning; consistent behaviors are summarized in Appendix~\ref{sec:app-other-models}. For all probing experiments, we fix prompts/seeds and capture per‑layer, per‑step activations and attention maps. 

\subsection{Repetitive Computation across Denoising Steps}

\label{sec:redundancy}



In causal video diffusion models, the historical context is typically attended at every denoising step. We investigate whether this repeated use of history is necessary in AR video generation.
Prior work on accelerated sampling suggests that useful structure can be established early in the denoising trajectory of image diffusion models~\cite{ma2024deepcache, selvaraju2024fora, liu2025taylorseer,zhang2025blockdance}. We identify a similar but distinct phenomenon in AR video diffusion models: middle-layer activations within the same frame show extremely high cosine similarity (above $0.95$), as illustrated in Fig.~\ref{fig:obs2_combined_motivational}. Given the high dimensionality of each feature vector (1536-d in Wan 2.1 1.3B~\cite{wan2025}), such consistently high cosine similarity indicates that the features are nearly identical across denoising steps. The effect is already visible at the earliest denoising steps, suggesting that the activations at early denoising steps stabilize quickly during the diffusion process.

The PCA visualization in Fig.~\ref{fig:obs2_combined_motivational} corroborates this finding: the principal components derived from the first denoising step closely align with those from later steps and successfully capture the object’s global shape, pose, and fine structural details (e.g., the curved, hook-like tail in the last frame) in the corresponding generated frame.
We attribute this pronounced feature similarity to the redundancy inherent in AR video generation—the features of the current frame are largely determined by historical contextual frames. Consequently, content and motion dynamics are effectively established in a single step, while subsequent denoising iterations primarily refine low-level pixel details and rendering quality (see Appendix Fig.~\ref{fig:layer-sim-combined} for extended analysis across layers).

\begin{figure}[ht] 
    \includegraphics[width=0.90\linewidth]{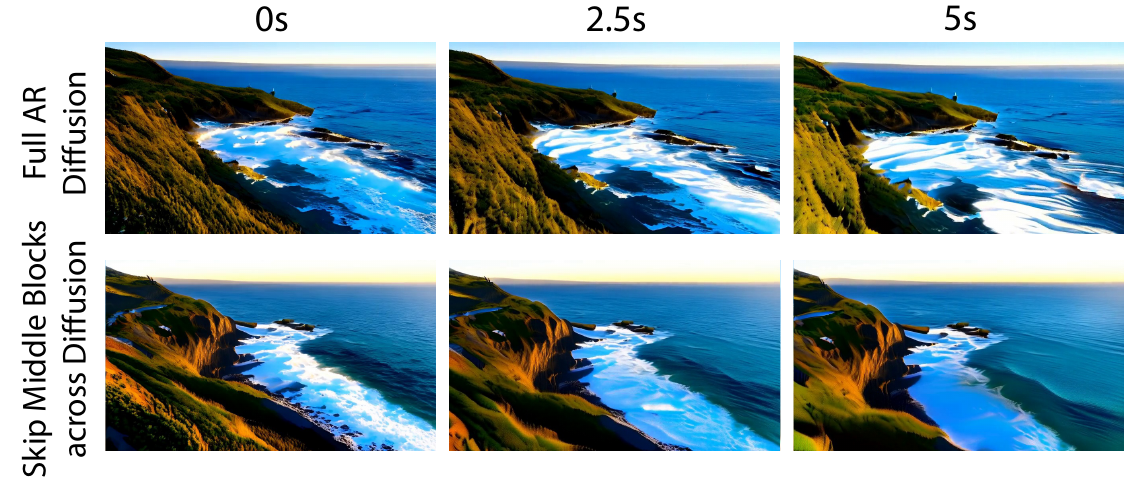}

    \caption{\textbf{Skipping the middle layers across denoising steps.}
    To take advantage of the repetitive computation, we finetune with a skip-layer design: except for the starting denoising steps, the denoiser skips a large chunk of 15 (out of 30) middle layers during diffusion. After short finetuning, semantics, layout, and motion are preserved and visual fidelity is restored. Full details on the design of this finetuning are provided in Appendix~\ref{sec:app-obs-redundancy}.}

    \label{fig:semantic_rendering_split}
    \vspace{-1em}
\end{figure}

To further verify the redundancy observation, we finetune the baseline with a skip-layer design (detailed in Fig.~\ref{fig:semantic_rendering_split} caption). Specifically, except for the first few denoising steps that run all 30 layers, subsequent steps skip layers 8--22 (15 middle layers), directly connecting early-layer outputs to late-layer inputs via residual connections.
As shown in Fig.~\ref{fig:semantic_rendering_split}, the skipped model successfully generates high-quality videos that faithfully preserve the object identities, spatial layout, and motion dynamics of the baseline model. This demonstrates that the new architecture does not learn a new generative manifold but instead operates within the same manifold as the baseline model.


\subsection{Deep Layers are Separable in Time}
\label{sec:sparsity}

\begin{figure}[ht] 
    \centering
    \includegraphics[width=\linewidth]{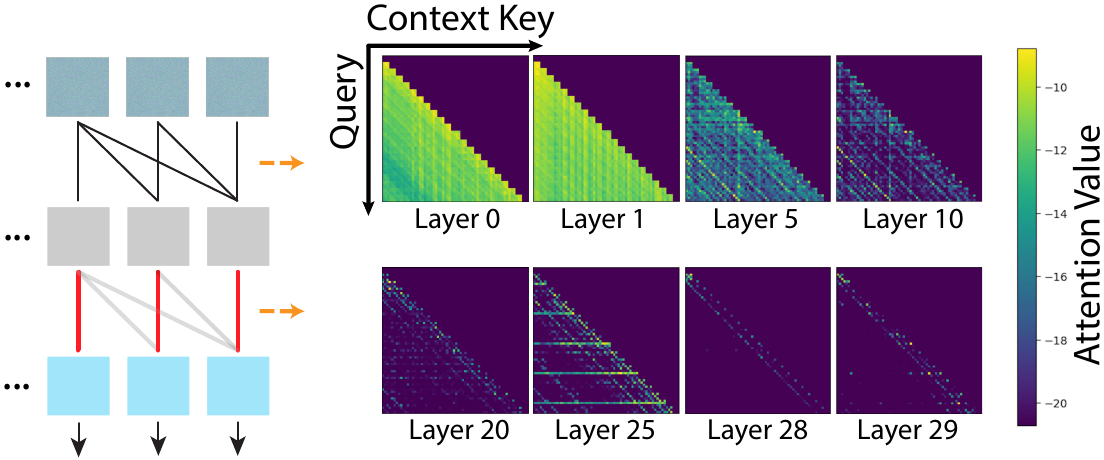}
    \caption{
    \textbf{Cross‑frame attention becomes sparse with depth.}
For a newly denoised frame $i$, we aggregate, for each transformer layer and attention head, the attention mass that query tokens at $i$ assign to keys from its context frames. Results indicate that deeper layers allocate markedly less mass to past frames, indicating they focus on intra‑frame refinement, and cross-frame attention is largely unnecessary.
    }
    \label{fig:sparse_attention}
    \vspace{-1em}
\end{figure}

To quantify how much temporal context is actually read at each layer, we compute the cross-frame attention mass across the AR video diffuser: for each transformer layer, we sum the attention from queries at frame $i$ to keys in frames $j<i$  (Fig.~\ref{fig:sparse_attention}).
This observation reveals a functional split in the model: early layers perform most temporal reasoning, while late layers focus on per‑frame rendering with little long‑range attention. Notably, although training uses a standard frame‑wise causal mask that permits dense cross‑frame attention, long‑range sparsity nonetheless emerges in deeper layers as an intrinsic property of the learned model.
\begin{figure}[t] 
    \includegraphics[width=0.95\linewidth]{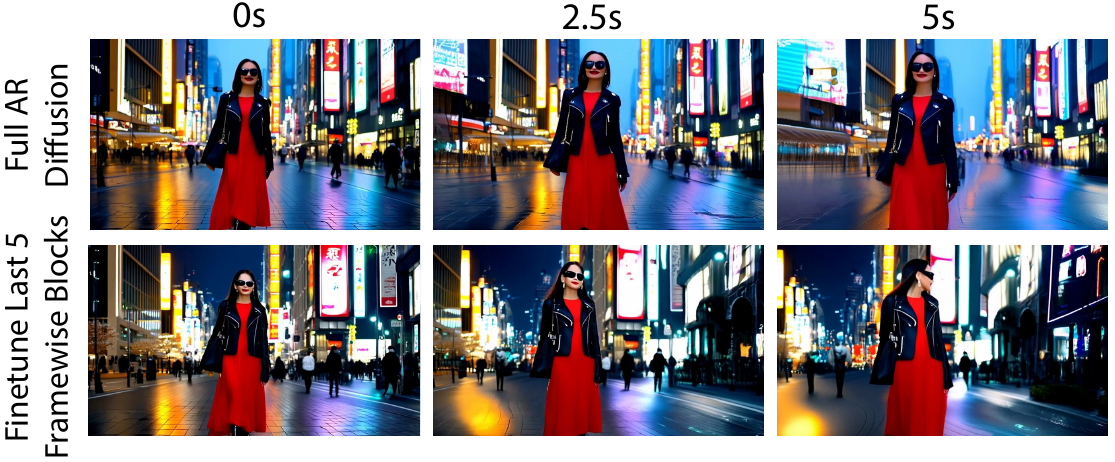}

    \caption{\textbf{Removing deep cross‑frame attention.}
  We switch the last $5$ (of $30$) layers from a frame‑causal mask to a frame‑diagonal mask, removing their access to context‑frame KV caches. A brief $5$k‑step finetune with the frame-diagonal mask stabilizes the semantics, layout, and motion and restores visual fidelity.}
    \label{fig:obs1_finetune_comparison}
    \vspace{-1em}
\end{figure}

Motivated by the observed long‑range sparsity, we investigate 
whether cross‑frame attention in deep layers can be removed in the architecture.
As shown in Fig.~\ref{fig:obs1_finetune_comparison}, a brief 5K-step finetuning on our partially frame-diagonal model effectively recovers the baseline visual generation quality.
We validate these observations on additional model families, including a 4-step block-autoregressive Self-Forcing model (Appendix Figs.~\ref{fig:sf-evidence},~\ref{fig:sf-attn}) and a 3D UNet trained with Diffusion Forcing (Appendix Fig.~\ref{fig:diffusion-forcing-appendix}), demonstrating that the separability patterns hold across different architectures and training objectives.

\section{\mymodel}
\label{sec:method}

The analyses of causal video diffusion models in \S\ref{sec:redundancy} and \S\ref{sec:sparsity} reveal two complementary regularities—step-wise invariance in early layers and temporal independence across frames in deeper layers. Together, these findings imply a functional separation within causal video diffusion models, whose operations consist of (1) producing and reasoning over clean context tokens, and (2) leveraging these context priors for iteratively denoising corrupted video-frame tokens. 
As a result, applying fully causal attention throughout the entire AR diffusion process leads to substantial redundant computation. Pruning these inactive paths naturally reduces both computational and memory overhead, motivating our efficient decoupled architecture, \textbf{\mymodel(\mymodelab)}, an encoder–decoder–style design that disentangles temporal causal reasoning from iterative denoising. Specifically, \mymodelab{} comprises a causal-reasoning encoder, which performs AR computations to produce context tokens without requiring iterative denoising, and a lightweight frame‑wise diffusion decoder, which focuses on synthesizing and refining the current frame conditioned on the context tokens from the encoder.

\subsection{Temporal Causal Encoder}

\label{sec:encoder}
Motivated by the step-wise redundancy observed in early layers (\S\ref{sec:redundancy}), we design a causal transformer encoder $\Encoder$ that runs once per generated frame, \emph{outside} the diffusion process, using causal attention over historical contexts stored as KV caches. Specifically, it computes a compact \textit{causal context} for the next frame:
\begin{equation}
    c_i \;=\; \Encoder \!\left(x_{<i},\, a_{\le i}\right),
\end{equation}
where $x_{<i}$ are the previously generated frames before time $i$, and $a_{\le i}$ denotes conditioning signals (e.g., actions). The context $c_i$ is a \emph{sequence} of latent tokens with the same spatial dimensions as the frame tokens (e.g., $H/p \times W/p$ tokens for patch size $p$), produced by the final layer of the encoder $\Encoder$. The causal context tokens $c_i$ summarize the history and are reused by the diffusion decoder across all denoising steps when generating the current video frame $x_i$ at time $i$.
Intuitively, $c_i$ encodes entities, layout, and motion cues anticipated for the generated frame at time $i$. Note that within $\Encoder$, attention among spatial tokens within each frame is bidirectional, whereas temporal attention across frames is causal.

\subsection{Frame‑wise Diffusion Decoder}
\label{sec:decoder}


Motivated by the cross-frame temporal independence observed in deep layers (\S\ref{sec:sparsity}), we introduce a \emph{lightweight frame‑wise diffusion decoder} $\Decoder$ that denoises noisy tokens corresponding to a video frame conditioned on the fixed contexts $c_i$ from $\Encoder$. In particular, $\Decoder$ learns to predict velocity $\hat{v}^t_i$ for a frame at time $i$ and denoising time step $t \!\in\!\{T,\dots,1\}$
\begin{equation}
\label{eq:dec}
\hat{v}^t_i \;=\; \Decoder \! \big(x_i^{t},\, t,\, c_i \big),
\end{equation}

The learned velocity $\hat{v}^t_i$ is used to iteratively denoise the corrupted video-frame tokens $x_i^{t}$ (starting from Gaussian noise) into a clean latent $x_i$. The context $c_i$ and noisy frame $x_i^t$ are combined via frame-wise token concatenation along the sequence dimension, forming a joint input sequence that the decoder processes with bidirectional self-attention. The diffusion decoder uses bidirectional attention \emph{within} each frame and does not propagate information \emph{across} frames; all historical context information is provided through the learned $c_i$ produced by the encoder $\Encoder$.

\begin{figure}[t]
    \centering
    \includegraphics[width=0.5\linewidth]{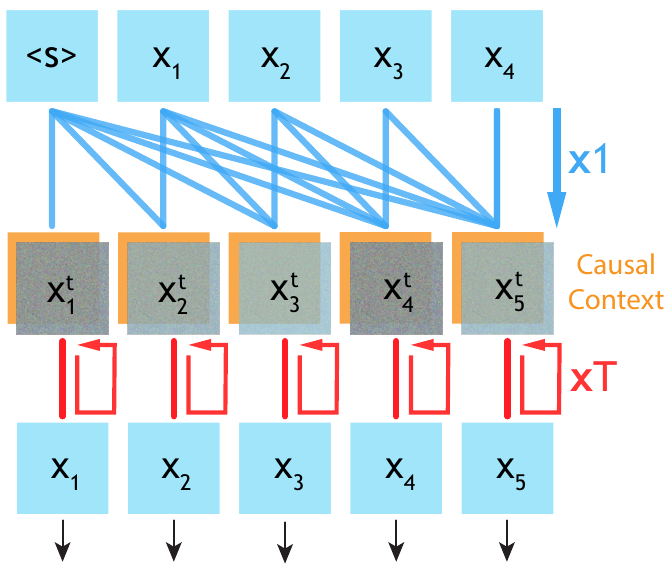}
    \caption{\textbf{Separable Causal Diffusion.} Once‑per‑frame causal reasoning produces a compact prior $c_i$, which the frame‑wise diffuser reuses across $T$ denoising steps to render $x_i$. }
    \label{fig:CSD}
    \vspace{-1em}
\end{figure}


\subsection{Training and Inference Framework} 
\label{sec:training_and_inference_framework}
\paragraph{Supervision.} 
Our encoder and decoder are trained jointly in an end-to-end manner, with the next-frame prediction objective (Fig.~\ref{fig:CSD}). In particular, the encoder $\Encoder$ takes in ground truth video frame tokens, processes them in parallel with causal attention, and generate a sequence of context tokens $\{ c_i \} _{i=1}^{N}$ corresponding to next frames, following Teacher Forcing training paradigm. The decoder $\Decoder$ then takes the noisy video-frame tokens $\{ x^{t}_i \}_{i=1}^{N}$ together with $\{ c_i \} _{i=1}^{N}$ to predict the corresponding velocities $\hat{v}^t_i$. The predicted velocities are supervised against the ground-truth conditional flow field (Equation~\ref{eq:cond_velocity}) using the loss defined in Equation~\ref{eq:loss}.
Since our diffusion decoder operates independently for each frame, we can improve token utilization during training by repeating each video-frame latent and sampling multiple noise scales per frame in the training sequence, similar to prior work~\cite{li2024mar, Zhang2025TestTimeTD}.

\noindent \textbf{Inference.} Because $\Encoder$ runs \emph{once per frame} and $\Decoder$ runs $T$ times within the frame, the amortized per‑frame time complexity is
\[
\underbrace{\mathcal{O}(\Encoder)}_{\text{once per frame}} \;+\; \underbrace{T\cdot \mathcal{O}(\Decoder)}_{\text{per denoising step}},
\]
with $\mathcal{O}(\Encoder) \gg \mathcal{O}(\Decoder)$ since $\Encoder$ performs inter‑frame causal attention with KV cache, whereas $\Decoder$ operates on each video frame latent independently .

Moreover, prior AR video diffusion models require an extra pass to cache the current generated frame content after its generation~\cite{yin2025causvid, huang2025selfforcing}. 
In contrast, as illustrated in Figure~\ref{fig:CSD}, our model follows a \emph{next-frame denoising} paradigm, in line with the design in autoregressive language models, which eliminates the need for an extra model invocation for KV caching. 

\paragraph{Corrupting causal context.}
To improve model robustness and context-following capability during inference, we inject noise to the historical context as done in prior work~\cite{chen2025skyreelsv2, song2025historyguidedvideodiffusion}. 
Because our architecture separates temporal reasoning from frame‑wise denoising, we can perturb their interface, the causal context $c_i$, as a means to inject corruption.
We adopt a simple Gaussian corruption,
\begin{equation}
\label{eq:causal-noise-operator}
\tilde{c}_i = c_i + \eta\,\boldsymbol{\zeta}, \qquad \boldsymbol{\zeta} \sim \mathcal{N}(0, I).
\end{equation}
Applying it during \emph{training}, it acts as an augmentation to reduce exposure bias. At \emph{inference}, it can also be used as a negative guidance signal. 
Compared with injecting noise to the frame tokens, corruption $c_i$ does not require extra passes of the network and is thus very efficient.
Empirically, we observe that modest noise corruption improves model robustness and context-following; full ablations are deferred to Appendix~\ref{sec:app-abl-corruption}.

\section{Experiments}
\label{sec:experiments}
\paragraph{Setting.} 
We evaluate our proposed \mymodel{} (\mymodelab{}) architecture in two complementary settings: 1) training from scratch on low-resolution video datasets, and 2) fine-tuning a high-resolution pretrained text-to-video diffuser to our architecture.
To study the effect of model capacity, we follow the DiT parameterization scheme (B/M/L)~\cite{peebles2023dit} and decompose the total transformer depth into a causal encoder and a diffusion decoder. We use the superscript $^E$ to denote variants with increased encoder depth and $^D$ to denote variants with increased decoder depth. Full hyperparameter specifications for all \mymodelab\ variants are provided in Appendix~\ref{sec:app-arch}. We additionally benchmark a fully causal video diffusion baseline trained through teacher-forcing, denoted as Causal-DiT.

\subsection{Training from Scratch on Small Video Datasets}
\begin{table}[t]
\centering
\footnotesize
\setlength{\tabcolsep}{4pt}
\renewcommand{\arraystretch}{1.1}
\caption{Comparison \& Ablation on \textbf{TECO--Minecraft 128${\times}$128}. 
}
\label{tab:main-mc}
\begin{tabular*}{\linewidth}{@{\extracolsep{\fill}}l c c c c c@{}}
\toprule
\multirow{2}{*}{Model} & \multirow{2}{*}{Sec/F} & \multicolumn{3}{c}{\textbf{144$\rightarrow$156}} & \multicolumn{1}{c}{\textbf{36$\rightarrow$264}} \\
\cmidrule(lr){3-5}\cmidrule(l){6-6}
 &  & LPIPS$\downarrow$ & SSIM$\uparrow$ & PSNR$\uparrow$ & FVD$\downarrow$ \\
\midrule
Latent FDM~\cite{ni2023lfdm}      & non-causal    & 0.429 & 0.349 & 13.4 & 167 \\
FAR-M-Long~\cite{gu2025far}       & 2.2           & 0.251 & 0.448 & 16.9 & 39  \\
Causal DiT-M                      & 2.4           & 0.196 & 0.512  & 18.9 & 38.7 \\
\midrule
\textbf{\mymodelab-M}             & \textbf{0.52} & 0.179  & 0.524  & 19.3 & 37.6 \\
\textbf{\mymodelab-M$^{E}$}& \textbf{0.52} & \underline{0.175} & \underline{0.525} & \underline{19.3} & \underline{36.1} \\
\textbf{\mymodelab-M$^{D}$}& 1.6          & \textbf{0.168} & \textbf{0.535} & \textbf{19.5} & \textbf{34.9}\\
\bottomrule
\end{tabular*}
\vspace{-1em}
\end{table}

\begin{table}[t]
\centering
\footnotesize
\setlength{\tabcolsep}{4pt}
\renewcommand{\arraystretch}{1.1}
\caption{Comparison \& Ablation on \textbf{UCF-101 64$\times$64}.}
\label{tab:main-ucf}
\begin{tabular*}{\linewidth}{@{\extracolsep{\fill}}l c c c c c@{}}
\toprule
\multirow{2}{*}{Model} & \multirow{2}{*}{Sec/F} & \multicolumn{4}{c}{\textbf{4$\rightarrow$12}} \\
\cmidrule(lr){3-6}
 &  & LPIPS$\downarrow$ & SSIM$\uparrow$ & PSNR$\uparrow$ & FVD$\downarrow$ \\
\midrule
RaMViD~\cite{hoppe2022ramvid}        & non-causal & 0.090 & 0.639 & 21.37 & 396.7 \\
MCVD\text{-}cp~\cite{voleti2022mcvd} & non-causal & 0.088 & 0.658 & 21.82 & 468.1 \\
FAR-B~\cite{gu2025far}               & 3.2        & 0.037 & 0.818 & 25.64 & 194.1 \\
Causal DiT-B                         & 3.9        & 0.038 & 0.827 & 25.85 & 187.6 \\
\midrule
\textbf{\mymodelab-B}                & \textbf{1.1}        & 0.038 & 0.824 & 25.78 & 174.7 \\
\textbf{\mymodelab-B$^E$}            & \textbf{1.1}        & \underline{0.037} & \textbf{0.829} & \underline{25.98} & \underline{171.1} \\
\textbf{\mymodelab-B$^D$}            & 2.8        & \textbf{0.036} & \textbf{0.829} & \textbf{26.00} & \textbf{158.7} \\
\bottomrule
\end{tabular*}
\vspace{-1em}
\end{table}

We evaluate pretraining performance on the widely used small-scale video generation benchmarks, TECO–Minecraft \cite{yan2023teco} and UCF-101 \cite{soomro2012ucf101}, as shown in Tables~\ref{tab:main-mc}~and~\ref{tab:main-ucf} (see also Appendix Table~\ref{tab:re10k-uncond} for RealEstate10K results).
We focus our comparison on diffusion-based methods, as non-diffusion baselines \cite{Babaeizadeh2021FitVid,Saxena2021Clockwork,hawthorne2022generalpurposelongcontextautoregressivemodeling} yield substantially poorer visual quality on these datasets.  
On TECO–Minecraft, \mymodelab-M attains the strongest overall generation quality—surpassing prior methods in LPIPS \cite{zhang2018lpips}, SSIM, PSNR, and FVD \cite{unterthiner2019fvd}—while delivering more than a 4× reduction in inference latency, measured as wall-clock seconds per frame (Sec/F, lower is better) on a single H100 GPU.
Similarly, on UCF-101 dataset, \mymodelab-B attains better or on par performance on all metrics and delivers more than 2× inference speedup relative to existing approaches.

Furthermore, we observe that having more layers for the causal encoder yields modest latency overhead but consistently improves generation quality, as shown by \mymodelab-M$^{\text{E}}$ in Table~\ref{tab:main-mc} and \mymodelab-B$^{\text{E}}$ in Table~\ref{tab:main-ucf} (see Appendix Table~\ref{tab:model-variants-decoupled} for detailed model configurations).
In addition, enlarging the diffusion decoder further boosts quality but incurs a substantial drop in inference speed. See \mymodelab-M$^{\text{D}}$ Table~\ref{tab:main-mc} and \mymodelab-B$^{\text{D}}$ in Table~\ref{tab:main-ucf}.

We further ablate the design choices of \mymodelab~and draw three conclusions. (1) Amortizing computation with a single once-per-frame encoder pass and multiple denoising passes significantly accelerates training (see Appendix Table~\ref{tab:amortization} and Fig.~\ref{fig:lpips-vs-time}).
(2) When providing the causal context $c_i$ and noisy frame $x_i$ to the diffusion decoder, frame‑wise concatenation consistently yields better performance than channel-wise concatenation (Appendix Table~\ref{tab:interface}).
(3) Injecting noise into the context via Eq.~\ref{eq:causal-noise-operator} improves robustness and generation quality (Appendix Table~\ref{tab:corruption+cfg}).
We refer readers to Appendix~\ref{sec:app-setup} and Appendix~\ref{sec:app-impl} for additional experimental details.

\begin{figure}[t]
    \centering
    \includegraphics[width=0.9\linewidth]{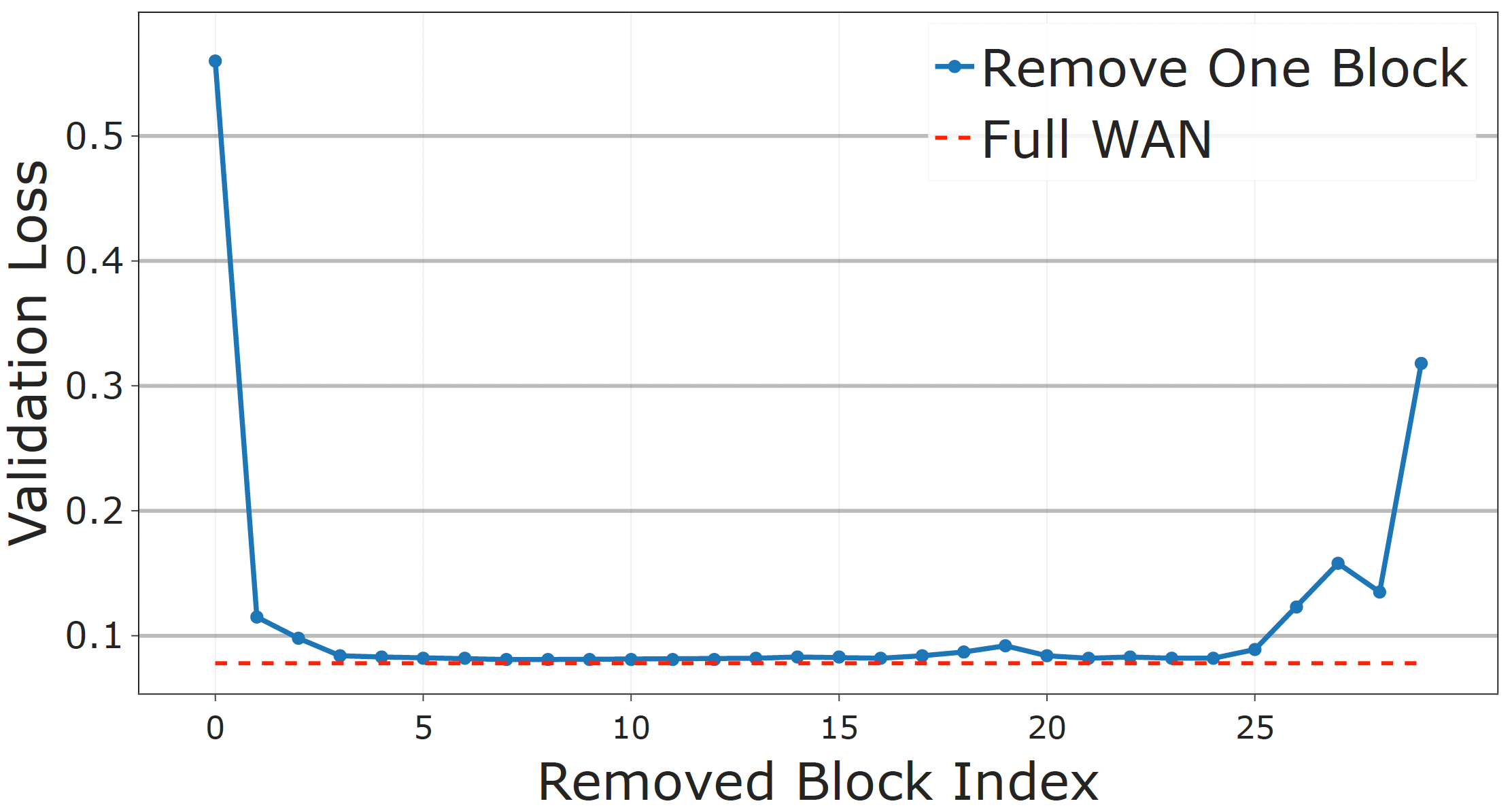}
    \caption{\textbf{The importance of transformer layers via leave-one-out inference.} We separately remove each layer in WAN2.1 T2V-1.3B and calculate the validation diffusion loss averaged across 5 noise levels. Results inform us which layers are important in finetuning.}
    \label{fig:leave_one_out_WAN}
\end{figure}

\subsection{Fine-Tuning Pretrained T2V Diffusion Model}
\label{sec:finetuning_experiments}

\paragraph{Architecture Adaptation.}
Recall from \S\ref{sec:training_and_inference_framework} that the causal encoder \mymodelab\ performs next-frame prediction by taking the previously generated frame $x_{i-1}$ at time $i-1$ as input to produce the context $c_i$ at time $i$.
In contrast, standard video diffusers require a noisy frame $x_i^{t}$ at time $i$ as input.
Empirically, we find that this architectural mismatch prevents reliable transfer of the capabilities learned by a pretrained T2V model to our \ModelNameShort\ architecture.

To bridge this gap, we align our causal encoder’s input distribution with that of the pretrained video diffusion model. Specifically, during training, we feed a corrupted current frame $x_i^{t}$ with high noise levels (top 20\%) into the encoder, whereas during inference we use pure Gaussian noise as the input.
This reparameterization preserves our decoupled design while matching the teacher’s input distribution, enabling stable and effective fine-tuning.

Moreover, we observe that the learned feature distributions across different layers of a pretrained video diffusion model~\cite{wan2025} differ substantially from the intended functionality of our \mymodelab\ design. As a result, a straightforward layer decomposition—treating early layers as the causal encoder and the remaining late layers as the diffusion decoder —introduces a large domain gap that undermines the pretrained model’s knowledge.
To identify which layers are essential for generation quality, we conduct a leave-one-out analysis shown in Fig.~\ref{fig:leave_one_out_WAN}. The results indicate that the earliest and latest layers contribute most to generation performance, whereas removing middle layers has much smaller negative impact. Motivated by this finding, we designate the first 25 layers of the pretrained 30-layer video diffusion model~\cite{wan2025} as the causal encoder, and combine its first 5 and last 5 layers to form the diffusion decoder, resulting in a total of 35 layers in our \mymodelab\ architecture (see Appendix Tables~\ref{tab:finetune-hparams} and~\ref{tab:sf-hparams} for detailed training hyperparameters).



\paragraph{Training and Distillation.} 

We apply the above architecture adaptation techniques to fine-tune our \mymodelab\ model using a conditional flow-matching loss and a teacher-forcing training strategy. To further obtain a few-step diffusion decoder, we adopt a self-forcing–style distillation approach~\cite{yin2024one, yin2024improved, huang2025selfforcing} and perform a full self-rollout of both the encoder and decoder, aligning the distribution of the generated samples with that of the pretrained bidirectional video diffusion teacher model. Additional training details are provided in Appendix~\ref{sec:app-finetune}.

\begin{figure}[t]
    \centering
    \includegraphics[width=0.9\linewidth]{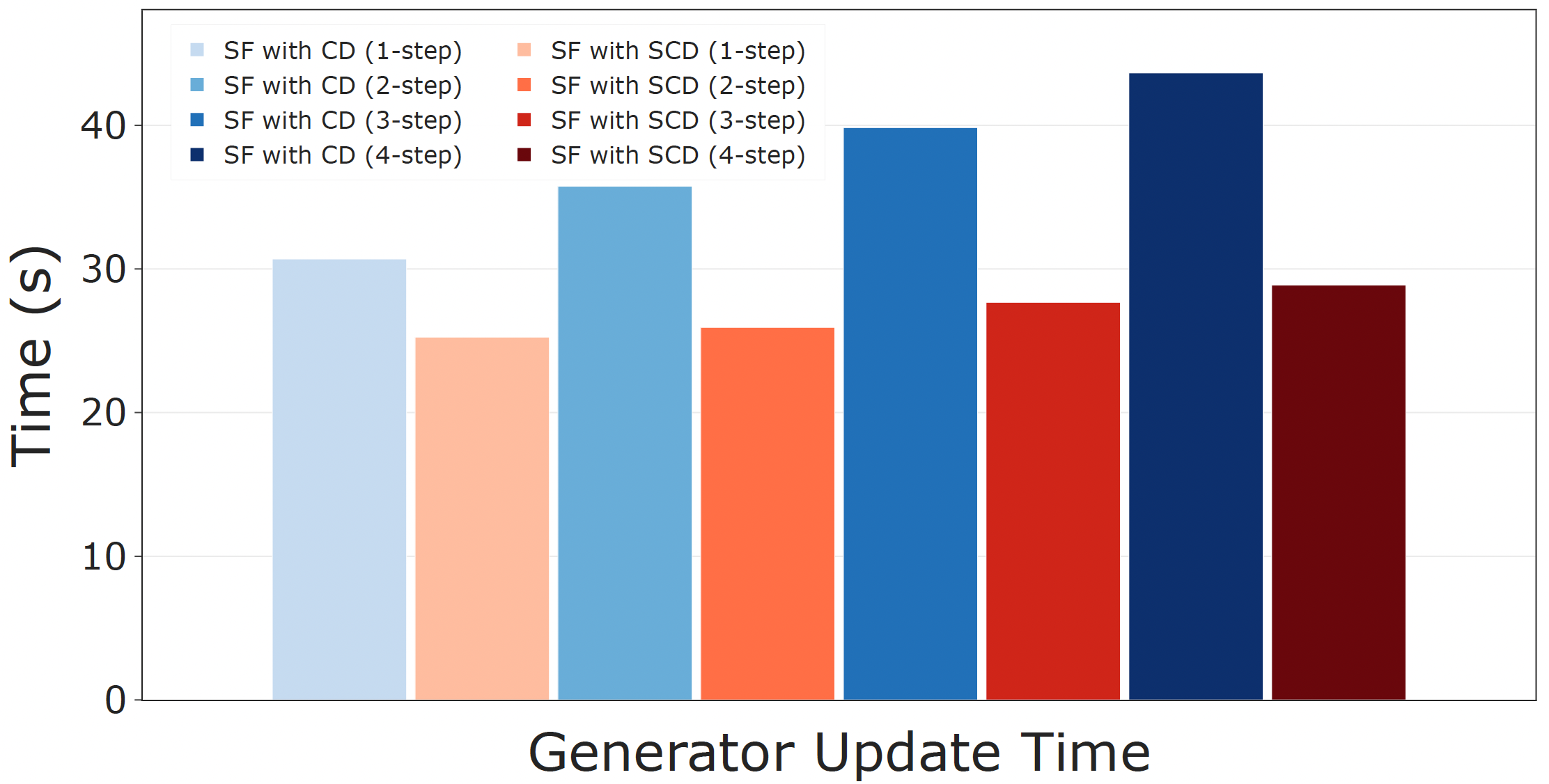}
    \caption{\textbf{Fine-tuning efficiency comparison.} Measured by per-iteration training time, SCD achieves superior training efficiency than causal diffusion baselines when performing frame‑wise sequential rollout distribution matching training.}
    \label{fig:sf_training_time}
    \vspace{-1em}
\end{figure}

\begin{table*}[t]
    \small
    \caption{\textbf{Text-to-Video quantitative comparison on VBench.} Models have similar parameter sizes and resolutions. Throughput (FPS)~$\uparrow$ and latency (s)~$\downarrow$ measured with batch size~1 on \textbf{1$\times$H100 80\,GB}. Higher is better for Total/Quality/Semantic scores~$\uparrow$.}
    \label{tab:main}
    \centering
    \resizebox{0.9\textwidth}{!}{
    \begin{tabular}{lccccccc}
        \toprule
        \multirow{2}{*}{Model} & \multirow{2}{*}{\#Params} & \multirow{2}{*}{Resolution} & \multirow{2}{*}{Throughput} & \multirow{2}{*}{Latency} &
        \multicolumn{3}{c}{Evaluation scores $\uparrow$} \\
        \cmidrule(lr){6-8}
         & & & (FPS) $\uparrow$ & (s) $\downarrow$ & Total & Quality & Semantic \\
         & & &                  &                  & Score & Score & Score \\
        \midrule
        \rowcolor{catgray}
        \multicolumn{8}{l}{\textit{Bidirectional diffusion models}}\\
        LTX-Video~\cite{hacohen2024ltx}      & 1.9B & $768{\times}512$ & 8.98          & 13.5          & 80.00 & 82.30 & 70.79 \\
        Wan2.1~\cite{wan2025}                & 1.3B & $832{\times}480$ & 0.78          & 103           & 84.26 & 85.30 & 80.09 \\
        \midrule
        \rowcolor{catgray}
        \multicolumn{8}{l}{\textit{Frame-wise autoregressive models}}\\
        NOVA~\cite{deng2024nova}             & 0.6B & $768{\times}480$ & 0.88          & 4.1           & 80.12 & 80.39 & 79.05 \\
        Pyramid Flow~\cite{jin2024pyramidal} & 2B   & $640{\times}384$ & 6.7           & 2.5           & 81.72 & 84.74 & 69.62 \\
        Self Forcing~\cite{huang2025selfforcing}     & 1.3B & $832{\times}480$ & \underline{8.9}  & \underline{0.45} & \textbf{84.26} & \textbf{85.25} & \textbf{80.30} \\
        \textbf{\mymodelab{} (Ours)}         & 1.6B & $832{\times}480$ & \textbf{11.1} & \textbf{0.29} & \underline{84.03} & \underline{85.14} & \underline{79.60} \\
        \bottomrule
    \end{tabular}
    }
    \vspace{-1em}
\end{table*}

\begin{figure*}[t]
    \centering
    \includegraphics[width=\textwidth]{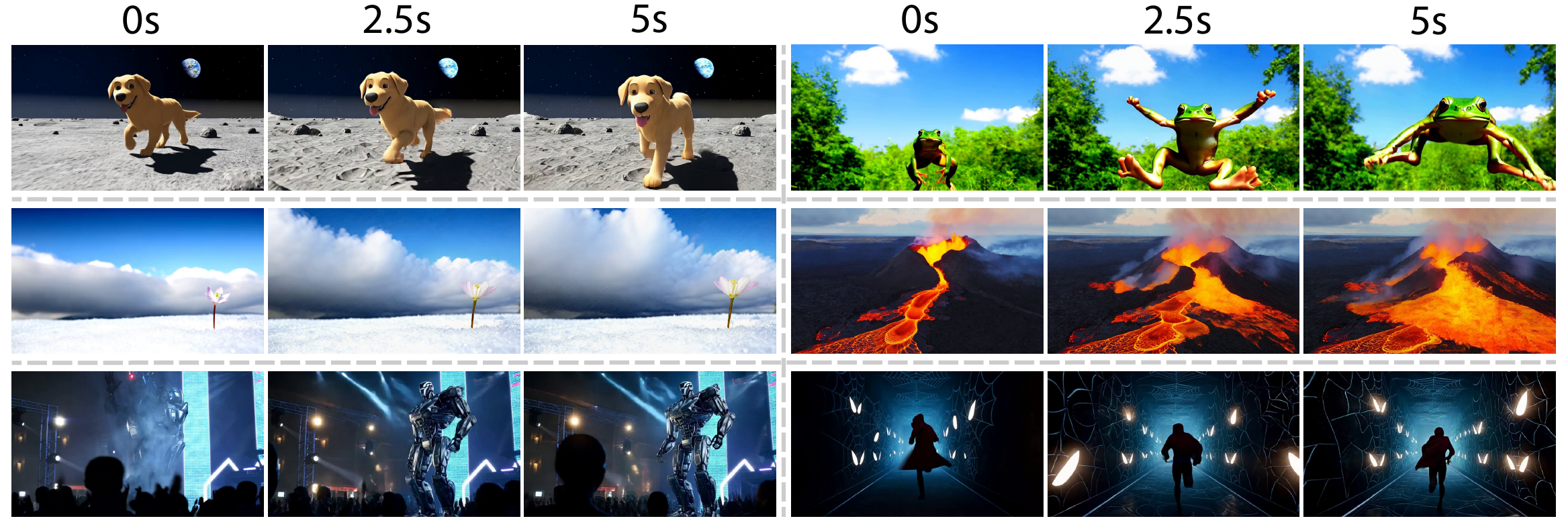}
    \caption{\textbf{Image-to-Video qualitative samples from \mymodelab{} fine-tuned from WAN 1.3B.} \mymodelab~preserves layout and motion while reducing per‑frame compute.}
    \label{fig:i2v_visualization}
    \vspace{-0.5em}
\end{figure*}

\paragraph{Results and analysis.}
Table~\ref{tab:main} reports text‑to‑video results on VBench \cite{huang2023vbenchcomprehensivebenchmarksuite}. Throughput and latency are measured on one H100 80\,GB GPU. 
Similar to the first-frame enhancement strategy of~\cite{yang2025onestepcausalvideogeneration}, we allocate extra compute to the initial frame and explicitly charge this overhead in throughput accounting.
Despite the architecture mismatch that typically penalizes quality when moving from a decoder-only architecture to a decoupled encoder–decoder, \mymodelab~(1.6B) achieves strong performance while being \textbf{$\sim$1.3$\times$ faster} than the frame‑wise Self Forcing baseline (11.1 vs.\ 8.9~FPS) with \textbf{$\sim$35\% lower latency} (0.29 vs.\ 0.45\,s). The total VBench score remains competitive (84.03 vs.\ 84.26), with similar quality and slightly lower semantic alignment, which we mainly attribute to the inevitable architectural mismatch. 
Compared to other AR baselines, \mymodelab~attains the highest throughput by a large margin (e.g., 11.1 vs.\ 6.7~FPS for Pyramid Flow), and is \emph{$>$10$\times$ faster} than a strong non‑causal diffusion model (Wan~2.1 at 0.78~FPS) while producing comparable overall scores. Figure~\ref{fig:i2v_visualization} shows qualitative \emph{I2V} examples from our finetuned model, where high visual quality and temporal consistency are observed with substantially lower inference cost.

Beyond inference efficiency, SCD enjoys high efficiency in rollout distribution matching training. As demonstrated in Figure~\ref{fig:sf_training_time}, SCD achieves 20\% higher training efficiency than Self Forcing in single-step rollout training, with marginal overhead in multi-step rollouts, indicating that SCD is more suitable for rollout training than full causal models.

\section{Conclusion}
\label{sec:conclusion}
Through probing and finetuning experiments, we identify two regularities in causal video diffusion models.
Firstly, 
middle-layer features of the denoiser exhibit strong consistency across denoising steps.
Secondly, cross-frame attention naturally becomes sparse with depth.
Building on these insights, we design \mymodel(\mymodelab{}), a novel architecture that decouples temporal reasoning from iterative denoising.
Across synthetic and real video benchmarks, \mymodel{} leads to substantial computational speedups while preserving generation quality.
Future work includes exploring the scaling law of next-frame denoising encoder against that of language models, exploiting the efficiency of SCD in roll-out based training frameworks, and integrating pre-trained reasoners and denoisers that lie in different latent spaces.

\paragraph{Limitations} 
Our decoupling assumes that temporal reasoning can be amortized across denoising steps and that deeper layers are predominantly intra‑frame—in densely pretrained models, both claims are approximations. 
(i) Step‑wise invariance weakens near the end of the trajectory: the similarity between middle-layer features in the last 10 denoising steps and the first 40 drops to about 0.8 (Fig.~\ref{fig:obs2_combined_motivational}), indicating that a single causal pass cannot fully substitute the evolving mid‑layer dynamics. 
(ii) Deep layers retain a small but non‑zero cross‑frame attention mass (Fig.~\ref{fig:sparse_attention}). 
These residual couplings plausibly account for the slight quality gap relative to fully causal‑attention baselines at high resolution (Table~\ref{tab:main}). 
Closing this gap may require more complicated architectural design to restore the missing dependencies while preserving the efficiency gain of separable temporal reasoning.

{
    \small
    \bibliographystyle{ieeenat_fullname}
    \bibliography{main}
}
\clearpage


\appendix


\newcommand{\Enc}{\mathcal{E}_\phi}
\newcommand{\Dec}{\mathcal{D}_\theta}
\newcommand{\ct}{c_t}
\newcommand{\Sden}{S} 
\newcommand{\lenc}{\ell} 
\newcommand{\mdec}{m}    

\section{Additional Analysis of Uncovering Causal Separability}
\label{sec:app-analysis}
\subsection{Redundancy across Denoising Steps}
\label{sec:app-obs-redundancy}
\begin{figure*}[t]
    \centering
    \begin{subfigure}{0.48\textwidth}
        \centering
        \includegraphics[width=\linewidth]{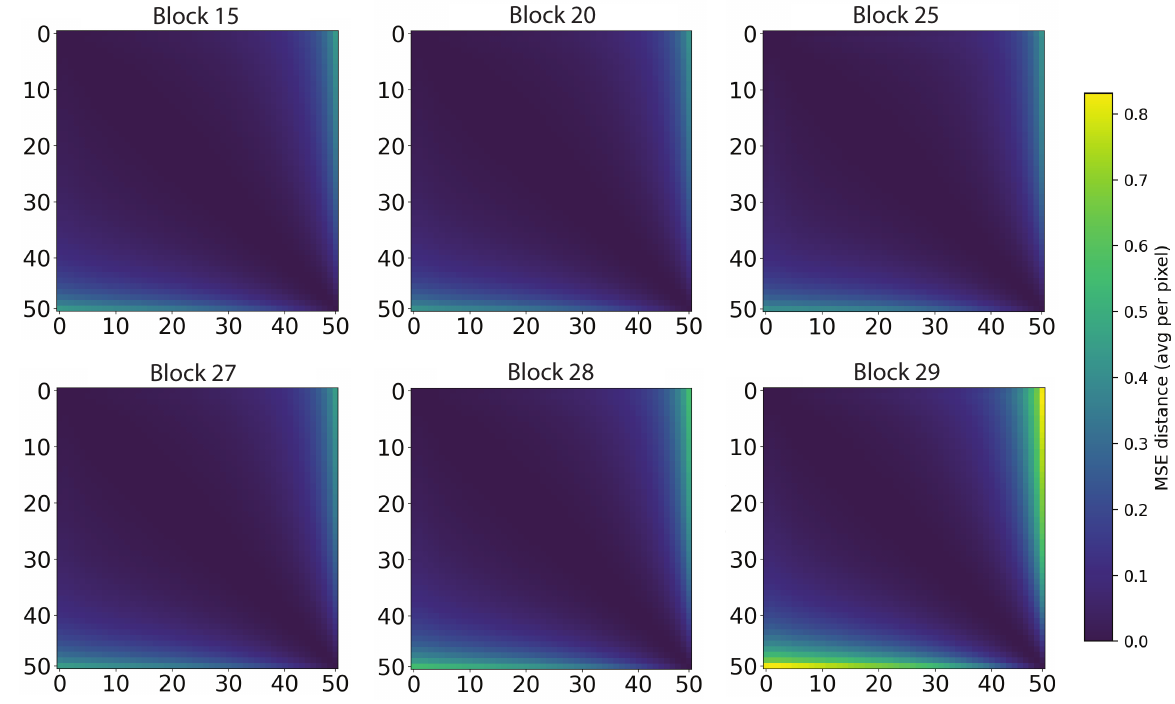}
        \caption{Step--step \emph{MSE-distance} matrices for multiple layers.}
        \label{fig:layer-sim-mats}
    \end{subfigure}
    \hfill
    \begin{subfigure}{0.48\textwidth}
        \centering
        \includegraphics[width=\linewidth]{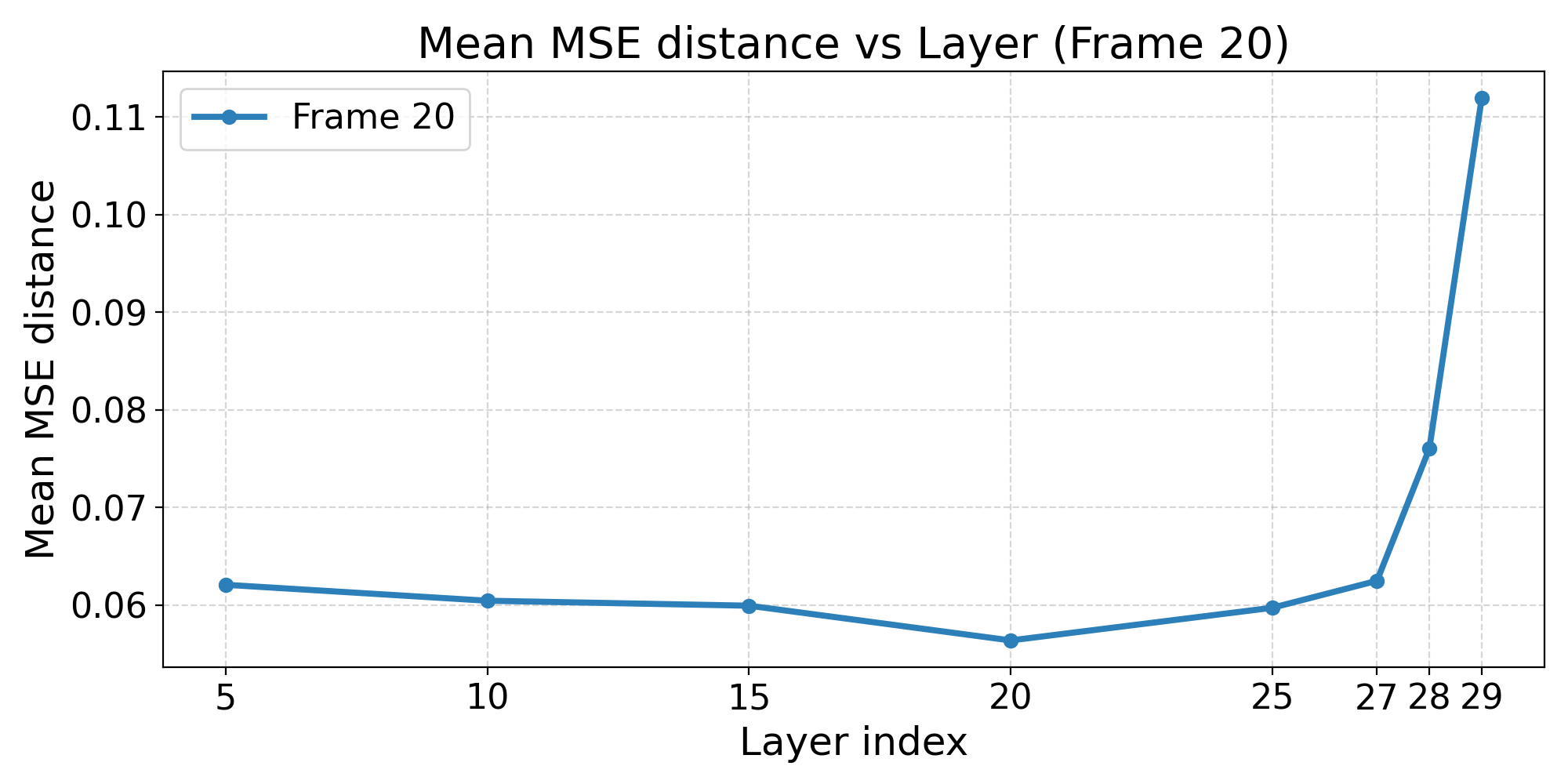}
        \caption{Mean MSE distance versus block index. Each value here represents the average value across the entire corresponding matrix on the left.}
        \label{fig:layer-sim-plots}
    \end{subfigure}
    \caption{\textbf{Redundant computation across denoising steps.}
    (a) Step--step feature \emph{MSE-distance} matrices of a fine-tuned AR WAN-2.1 T2V-1.3B model at several layer depths
    (layers 5, 10, 15, 20, 25, 28, 29). Middle layers (10--25) show broad, \emph{low-distance} bands across all 50 denoising steps, indicating that their features are mostly invariant along the diffusion trajectory, whereas the last few layers
    exhibit more step-dependent distances. (b) A complementary per-layer summary: the average MSE distance across all pairs of denoising steps remains small in the early and middle layers, but increases dramatically in late layers. Together, these views support our claim that early/middle blocks perform largely redundant
    computation across denoising steps, while the deepest blocks remain step-specific for intra-frame rendering, motivating
    our design that amortizes the first 25 layers once per frame and reuses them across all denoising steps. 
    }
    \label{fig:layer-sim-combined}
\end{figure*}

\paragraph{Feature similarity across denoising steps and depth.}
We extend the analysis in Fig.~\ref{fig:obs2_combined_motivational} to multiple depths of an autoregressively fine‑tuned WAN‑2.1 T2V‑1.3B model. For a fixed set of prompts/seeds, we roll out the model for 50 denoising steps and, at every transformer block $\ell$ and step $s$, record the hidden features. We then compute a step–step \emph{mean‑squared‑error (MSE) distance} matrix
$\mathbf{S}_\ell \in \mathbb{R}^{T\times T}$ with entries
$[\mathbf{S}_\ell]_{s,s'}=\|f_{\ell,s}-f_{\ell,s'}\|_2^2$, where $f_{\ell,s}$ denotes the layer‑$\ell$ features at step $s$. Representative matrices for $\ell\in\{10,15,20,25,28,29\}$ are shown in Fig.~\ref{fig:layer-sim-mats} and Fig.~\ref{fig:layer-sim-plots} 

Layers $10$–$25$ exhibit pronounced step‑wise invariance: their similarity maps contain broad, near‑uniform high‑value bands (also visible at the 15th layer in the main paper), indicating that middle/early denoiser blocks repeatedly recompute almost the same features across the diffusion trajectory. In contrast, the last two layers ($\ell=28,29$) display markedly lower and more step‑dependent similarity, consistent with these blocks performing step‑specific, intra‑frame rendering. 

To further verify the redundancy observation, we fine-tune the baseline with a skip-layer design, in which the majority of denoising steps skip the middle-layers computation. 

To further verify the redundancy finding, we fine‑tune the causal baseline with a \emph{skip‑layer} schedule in which most denoising steps bypass the middle of the network. Concretely, only the first five denoising steps run the full denoiser; all subsequent steps traverse just a short \emph{prefix} of 5 early layers and a \emph{suffix} of 10 late layers, skipping the middle chunk. We retain the prefix because, as shown in Fig.~\ref{fig:leave_one_out_WAN} in the main text, early layers are particularly important during fine‑tuning. The outcome (Fig.~\ref{fig:semantic_rendering_split} in the main text) is that the skipped model produces high‑quality videos and recovers the visual fidelity of the fully causal baseline, validating that most cross‑step computation in early/middle layers is redundant and can be shared. This experiment is designed to test the observation in the simplest setting. In later SCD fine‑tuning, we adopt a more aggressive recipe that also works: the first 25 layers run \emph{once per frame} (amortized across steps), and only the final $5{+}5$ layers participate in per‑step denoising under our SCD design.

\subsection{Evidence on Other Models} 
\label{sec:app-other-models}


Beyond the WAN‑2.1 (1.3B) model, we also evaluate an open‑source, open‑weight \textbf{Self‑Forcing} (SF) 1.3B model~\cite{huang2025selfforcing}—a few‑step, \emph{block‑autoregressive} student distilled from a bidirectional teacher that predicts three latent frames per block. We chose this model deliberately for two reasons. \textbf{(i) Block‑autoregression} is a widely used generative pattern in contemporary video systems, and even in emerging language diffusion models~\cite{arriola2025block}, so validating our analysis on a block‑AR student makes the conclusions relevant to a broad class of architectures~\cite{lin2025aapt,yang2025longlive,shin2025motionstream}. \textbf{(ii) Self‑forcing} is the prevailing post‑training recipe for large autoregressive video models, bridging the train–test gap and distilling many‑step teachers into efficient few‑step samplers~\cite{huang2025selfforcing,cui2025selfforcingpp,liu2025rollingforcing,fastvideo2025causalwan22}. By observing our claims on both a many‑step WAN and a few‑step, block‑autoregressive, self‑forced student, we cover complementary ends of the design space; the aligned observations across these regimes would substantially strengthen the generality of our claims.

\begin{figure*}[t]
    \centering
    \begin{subfigure}{0.98\textwidth}
        \centering
        \includegraphics[width=\linewidth]{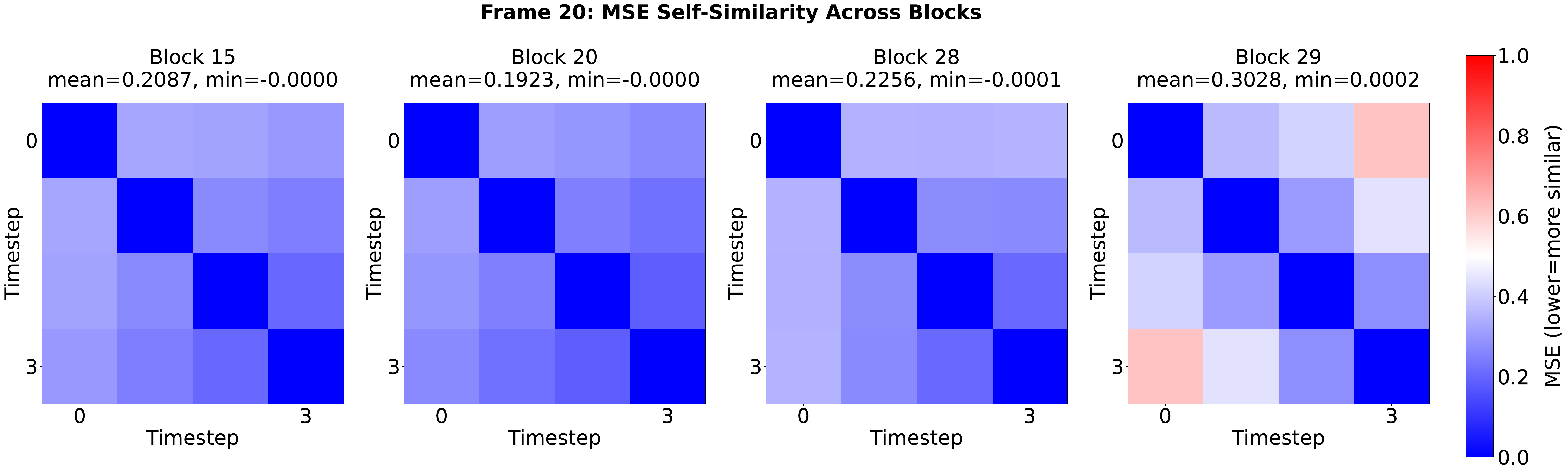}
        \caption{\textbf{Self‑forcing: step–step feature similarity} across multiple layers (analogous to Fig.~\ref{fig:obs2_combined_motivational}(a) in \S\ref{sec:analyses}). Early/middle layers show broad high‑similarity (small MSE) bands over denoising steps.}
        \label{fig:sf-sim}
    \end{subfigure}

    \begin{subfigure}{0.98\textwidth}
        \centering
        \includegraphics[width=\linewidth]{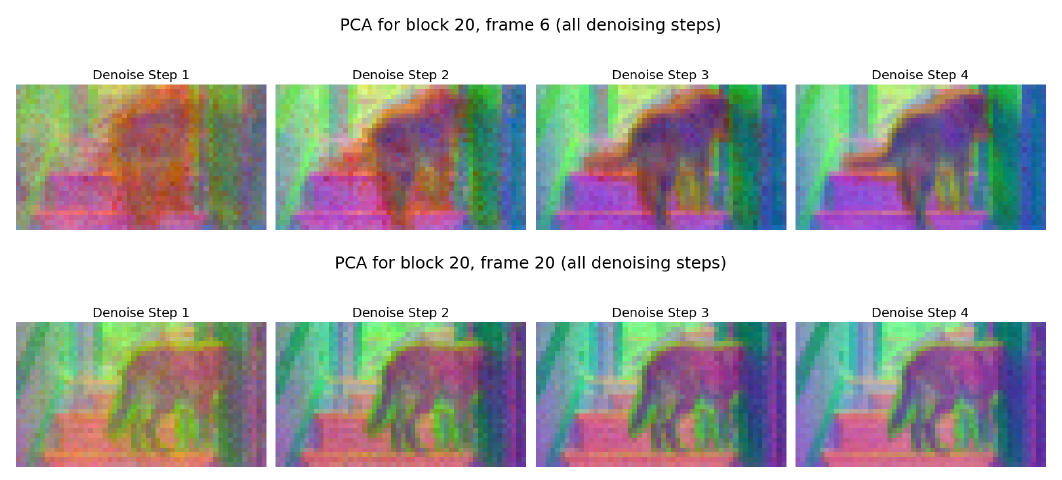}
        \caption{\textbf{Self‑forcing: PCA of activations} (combined view; analogous to Fig.~\ref{fig:obs2_combined_motivational}(b) in \S\ref{sec:analyses}). Principal components remain stable across denoising steps and across \emph{blocks} in the block‑autoregressive rollout.}
        \label{fig:sf-pca}
    \end{subfigure}
    \caption{\textbf{Evidence on another model family (self‑forcing, 4‑step, block‑autoregressive).}
    We replicate the \S\ref{sec:analyses} analyses on a few‑step self‑forcing student.
    (a) Early/middle layers exhibit high step–step feature similarity.
    (b) PCA views confirm that principal directions stabilize early and change little across denoising steps and \emph{blocks}.}
    \label{fig:sf-evidence}
\end{figure*}

\begin{figure*}[t]
    \centering
    \includegraphics[width=\linewidth]{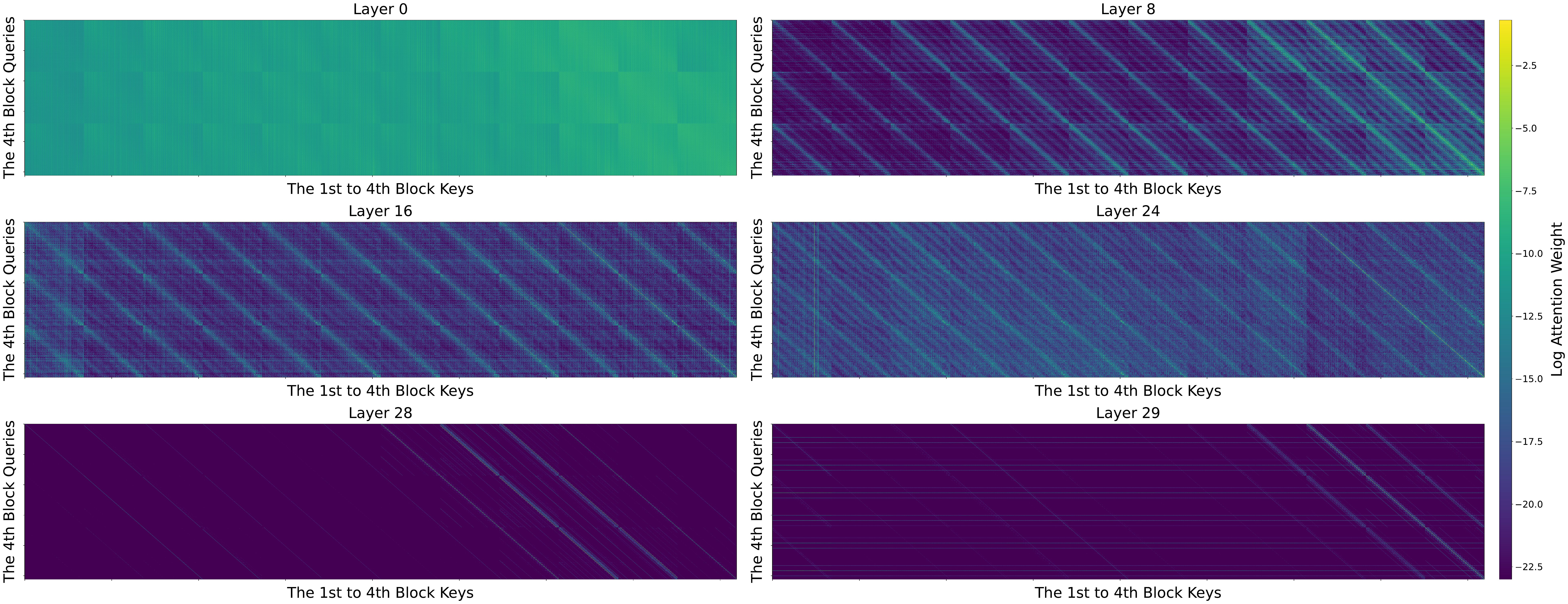}
    \caption{\textbf{Evidence on another model family (self‑forcing, 4‑step, block‑autoregressive).}
    We replicate the \S\ref{sec:analyses} analyses on a few‑step self‑forcing student. As with the multi-step models, its deeper layers place minimal attention on past blocks, again revealing strong temporal sparsity.}
    \label{fig:sf-attn}
\end{figure*}

\paragraph{Observations.}
We see the same patterns as in \S\ref{sec:analyses}. Early and middle \emph{layers} produce very similar features across denoising steps. PCA views change little with the step index and already capture global structure in the \emph{first block}. Deeper layers are temporally sparse and focus on intra‑frame rendering. These results indicate that the separability trends hold for both a 50‑step WAN and a distilled 4‑step \emph{block‑autoregressive} self‑forcing model.

\begin{figure*}[t]
    \centering 
    \begin{subfigure}[t]{0.48\textwidth}
        \includegraphics[width=\textwidth]{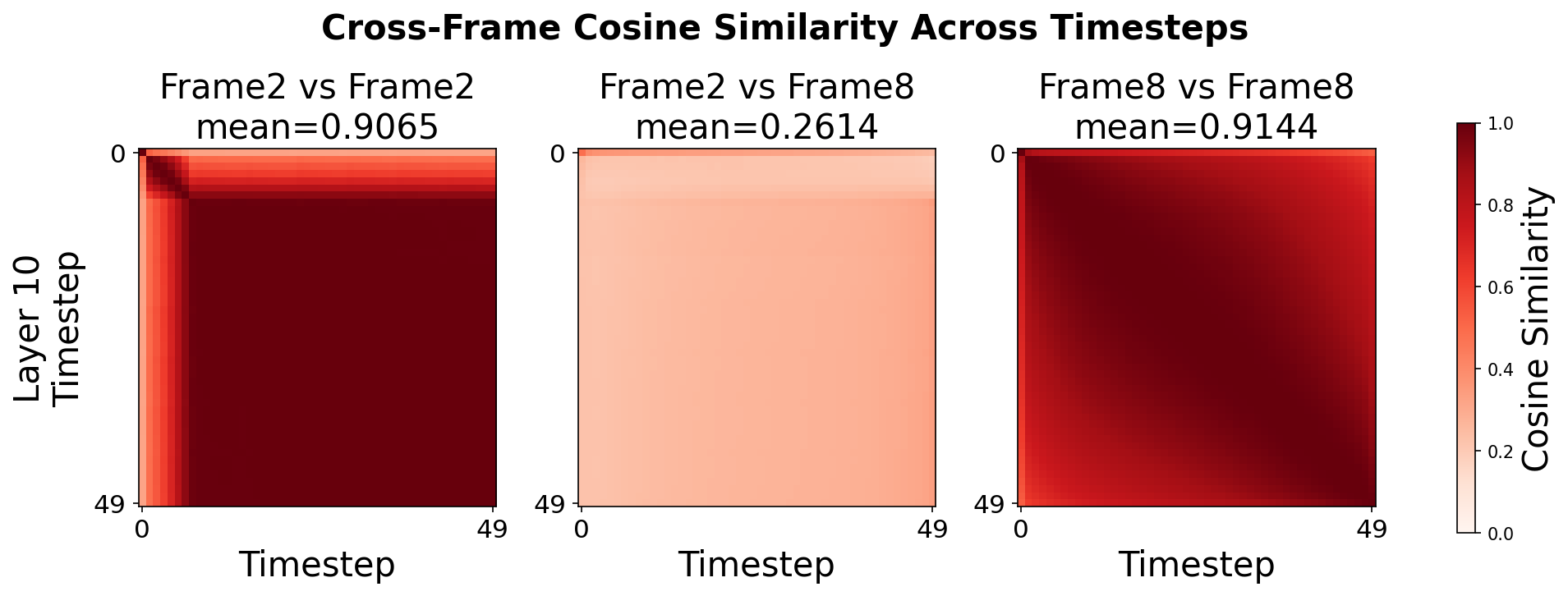}
        \caption{Cross-step cosine similarity}
    \end{subfigure}
    \hfill
    \begin{subfigure}[t]{0.48\textwidth}
        \includegraphics[width=\textwidth]{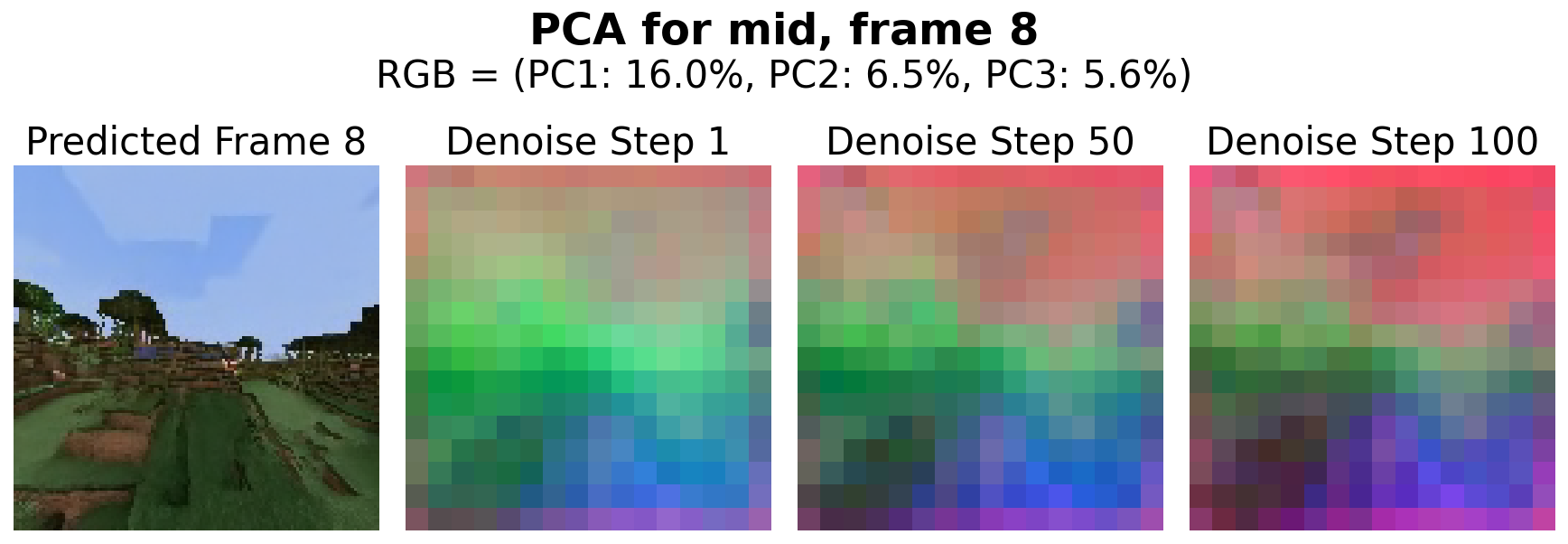}
        \caption{PCA subspace alignment}
    \end{subfigure}

    \vspace{14pt}

    \begin{subfigure}[t]{0.7\textwidth}
        \centering
        \includegraphics[width=\textwidth]{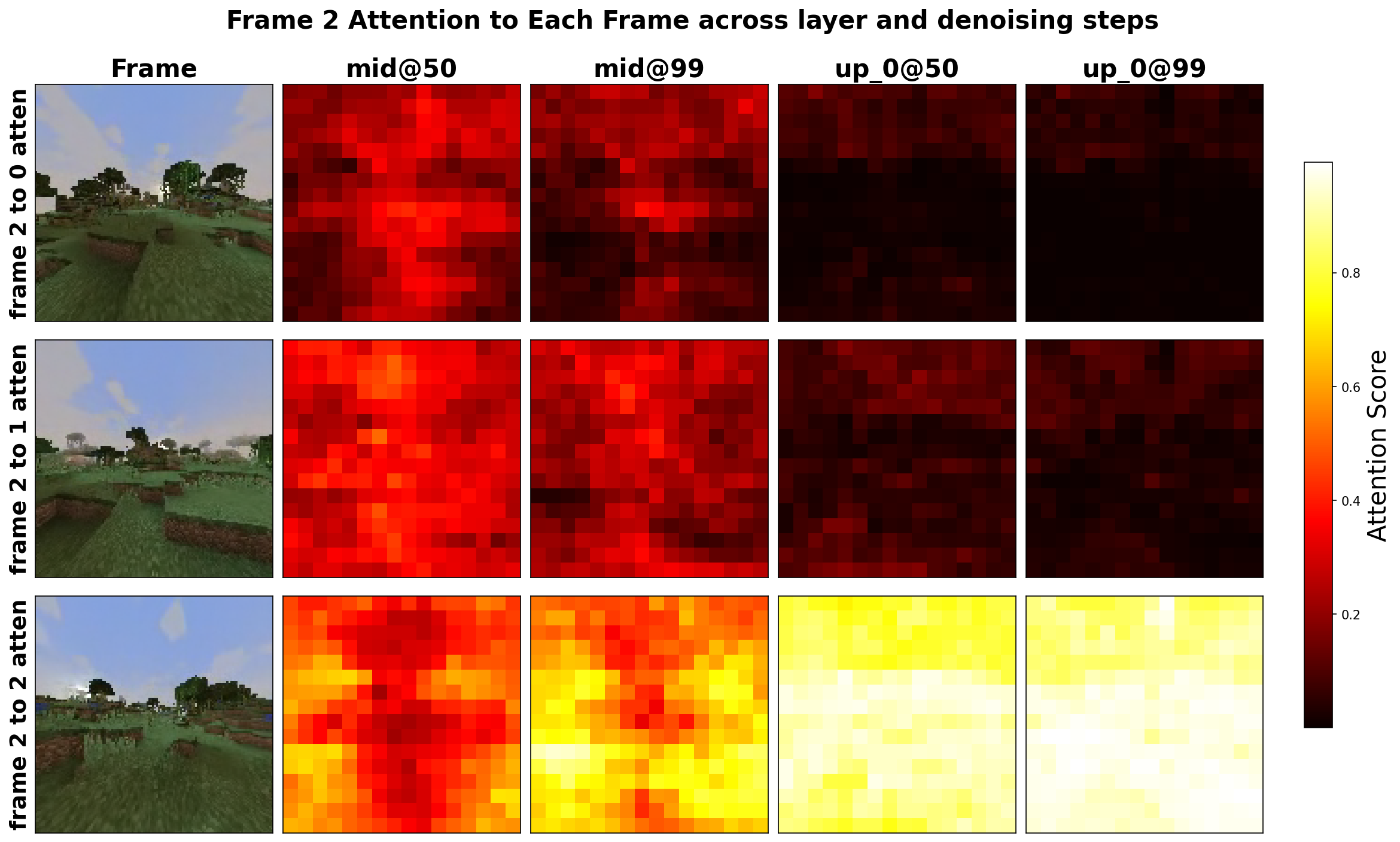}
        \caption{Attention visualization}
    \end{subfigure}
    
    \caption{\textbf{Generality to a 3D UNet on Minecraft under Diffusion Forcing.} Despite the substantially different backbone (UNet vs.\ Transformer) and training objective (Diffusion Forcing vs.\ Teacher Forcing), we observe the same qualitative trends: mid-layer representations stabilize early in denoising, exhibiting (a) high cross-step cosine similarity and (b) strong PCA subspace alignment. Moreover, (c) deep denoiser layers attend sparsely to the context frames (rows 1–2) while focusing primarily on intra-frame structure (row 3). We demonstrate this phenomenon is consistent across denoising timesteps (e.g., $t{=}50$ vs.\ $t{=}99$).}
    \label{fig:diffusion-forcing-appendix}
\end{figure*}

\paragraph{Evidence on Diffusion Forcing with 3D UNet.}
To further validate the generality of our observations beyond Transformer-based architectures and teacher-forcing training, we analyze a 3D UNet trained with Diffusion Forcing~\cite{chen2024diffusionforcing} on Minecraft. Despite the substantially different backbone (UNet vs.\ Transformer) and training objective (Diffusion Forcing vs.\ Teacher Forcing), we observe the same qualitative trends as shown in Fig.~\ref{fig:diffusion-forcing-appendix}. Mid-layer representations stabilize early in denoising, exhibiting high cross-step cosine similarity and strong PCA subspace alignment. Moreover, deep denoiser layers attend sparsely to the context frames while focusing primarily on intra-frame structure. We also demonstrate this phenomenon is consistent across denoising timesteps (e.g., $t{=}50$ vs.\ $t{=}99$). These results provide strong evidence that the causal separability we observe is a fundamental property of causal video diffusion rather than model-specific artifacts.

\section{Ablation Studies}
\label{sec:app-ablations}

\subsection{Encoder-Decoder Interface}
\label{sec:app-abl-interface}
We compare two ways of providing the context latent $\ct$ to the frame‑wise diffusion decoder $\Dec$:
(i) \emph{Channel Concatenation} we concatenate $\ct$ with the noisy frame tokens along the channel dimension, project back to the standard channel dimension with a linear layer, and feed it to the decoder as input;
(ii) \emph{Frame Concatenation} we prepend $\ct$ as a prefix frame for the noisy frame, and then feed them into the decoder. 
This effectively positions $\ct$ as the context frame of the current frame, performing self-attention together as a whole sequence. We also ablate the positional embedding applied to the context frame, either embedding it as "the last temporal frame" or "the current frame".
As shown in Tab.~\ref{tab:interface}, sequence fusion with positional embedding as a historical frame outperforms the alternatives.

\begin{table}[t]
\centering
\small
\setlength{\tabcolsep}{8pt}
\caption{\textbf{Ablations on the encoder-decoder interface.} Sequence fusion outperforms channel fusion; temporal RoPE is slightly better than identical (non-causal) RoPE. For simplicity of ablation, metrics are reported at 400k training steps, with the unified "144 context frames, 156 generated frames" evaluation setup.}
\label{tab:interface}
\begin{tabular}{lcc}
\toprule
Encoder-Decoder Interface & FVD $\downarrow$ & LPIPS $\downarrow$ \\
\midrule
Channel dim.                      & 25.4 & 0.231 \\
\textbf{Frame dim. with temporal RoPE}& \textbf{24.8} & \textbf{0.219} \\
Frame dim. with identical RoPE    & 25.1 & 0.223 \\
\bottomrule
\end{tabular}
\end{table}


\subsection{Training Noisy Batches: Amortized Multi‑Sample Decoding}
\label{sec:app-abl-amortized}
Because $\Enc$ consumes only clean history, it is noise‑agnostic. We therefore perform one efficiency trick in training: for each batch of clean videos, we encode once per frame to obtain $\ct$, then draw $K$ i.i.d.\ noise/timestep pairs for the current frame and run $\Dec$ $K$ times, saving the amortized cost for learning each noisy batch.
As Tab.~\ref{tab:amortization} demonstrates, throughput of noisy batch increases with $K$, which also improves the training speed.

\begin{table}[t]
\centering
\footnotesize
\setlength{\tabcolsep}{6pt}
\caption{\textbf{Amortized multi‑sample decoding in training.} Encoder depth $8$, decoder depth $4$, as in our SCD-B and SCD-M models. For simplicity of ablation, metrics are reported at 400k training steps, with the unified "144 context frames, 156 generated frames" evaluation setup.}
\label{tab:amortization}
\begin{tabular}{ccccc}
\toprule
$K$ & BP/clean batch & BP/noisy batch & noisy batch/s  & FVD \\
\midrule
1 & $8 + 4{\times}1 = 12$ & $8/1 + 4 = 12$ & \textit{22.0} & \textit{23.9} \\
2 & $8 + 4{\times}2 = 16$ & $8/2 + 4 = 8$  & \textit{39.2} & \textit{23.3} \\
4 & $8 + 4{\times}4 = 24$ & $8/4 + 4 = 6$  & \textit{63.0} & \textit{\textbf{23.1}} \\
\bottomrule
\end{tabular}\\[2pt]

\end{table}

\begin{figure*}[t]
    \centering
    \includegraphics[width=0.7  \linewidth]{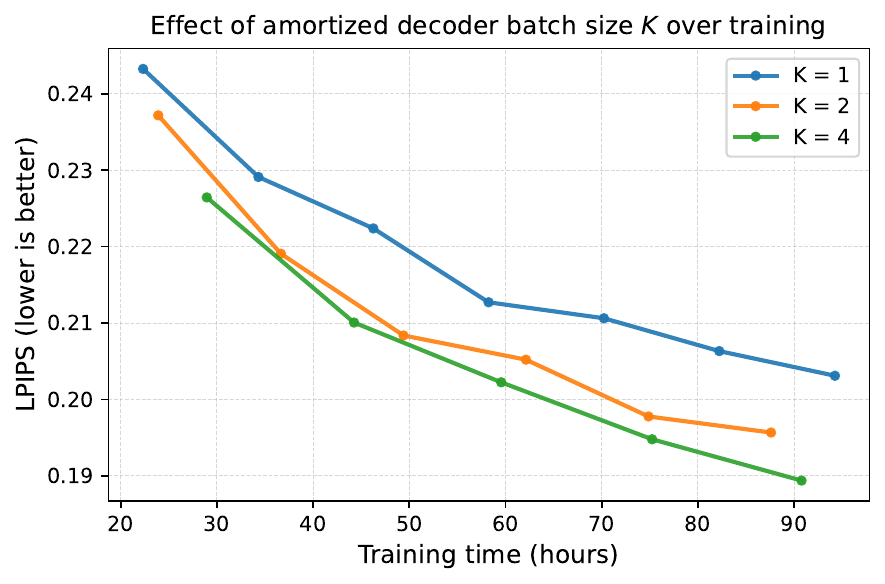}
    \caption{\textbf{Effect of amortized decoder batch size $K$ at matched training time.}
    LPIPS on the validation set versus wall-clock training time (hours) for $K\!\in\!\{1,2,4\}$.
    Even when comparing at equal training time rather than equal optimization steps, larger
    $K$ achieves lower LPIPS, indicating genuine gains from amortizing multiple noisy
    decoder samples per encoder pass.}
    \label{fig:lpips-vs-time}
\end{figure*}

\paragraph{Training-time comparison.}
The main-table ablation in Sec.~\ref{sec:app-abl-amortized} compares different $K$ at matched optimization steps, which slightly favors larger $K$ because each step processes more independently noised targets.
To control for this, Fig.~\ref{fig:lpips-vs-time} re-plots LPIPS against \emph{wall-clock training time}.
Despite the heavier per-step compute, curves with $K{=}2$ and especially $K{=}4$ reach lower LPIPS than $K{=}1$ at the same elapsed time.
This shows that amortized multi-sample decoding does more than just see more noise per step: reusing the once-per-frame encoder output across multiple noisy decoder calls yields a better optimization trajectory even under a fixed compute budget.

\subsection{Noisy Context Latent: Context Corruption and CFG}
\label{sec:app-abl-corruption}
We perturb only the context $c_i$ by adding Gaussian noise
$\tilde c_t = c_t + \eta_{\mathrm{t}}\epsilon$, $\epsilon \sim \mathcal{N}(0,I)$,
where $c_t$ is normalized to unit variance and $\eta_{\mathrm{t}}^2$ is the noise variance.
Table~\ref{tab:corruption+cfg} reports two complementary ablations on TECO--Minecraft at 400k steps.
On the \emph{left}, we vary the training-time corruption strength
$\eta_{\mathrm{t}} \in \{0, 0.05, 0.10, 0.20, 0.50\}$ and evaluate without any
inference-time classifier-free guidance (CFG) on $c_t$:
moderate noise around $\eta_{\mathrm{t}}\!=\!0.05$ improves FVD while stronger corruption eventually hurts.
On the \emph{right}, we fix training-time corruption at $\eta_{\mathrm{t}}{=}0.05$ and ablate CFG using a negative branch with the same perturbation
$\tilde c_t = c_t + 0.05\,\epsilon$ and guidance scale
$\eta_{\mathrm{cfg}} \in \{0.0, 1.0, 1.5, 2.0\}$; $\eta_{\mathrm{cfg}}{=}1.5$ gives the best trade-off.
Overall, modest training-time corruption plus a small CFG weight on the corrupted causal prior
($\eta_{\mathrm{t}}\!\approx\!0.05$, $\eta_{\mathrm{cfg}}\!\approx\!1.5$) yields the strongest long-horizon visual quality.
Because these perturbations act only on the causal interface, they do not require re-caching any context frames.

\begin{table}[t]
\centering
\footnotesize
\setlength{\tabcolsep}{6pt}
\caption{\textbf{Training-time causal corruption and test-time CFG with corruption (Minecraft, 400k steps).} Left: vary $\eta_{\mathrm{t}}$ at training and evaluate with no CFG on $c_t$ ($\eta_{\mathrm{cfg}}{=}0$). Right: fix $\eta_{\mathrm{t}}{=}0.05$ at training and sweep inference-time CFG scale $\eta_{\mathrm{cfg}}$ on a corrupted causal prior $\tilde c_t = c_t + 0.05\,\epsilon$.}
\label{tab:corruption+cfg}
\begin{tabular}{lcc|lcc}
\toprule
\multicolumn{3}{c|}{\textbf{Training corruption}} & \multicolumn{3}{c}{\textbf{Inference CFG with corruption} }\\
\cmidrule(lr){1-3} \cmidrule(lr){4-6}
Noise $\eta_{\text{t}}$ & FVD $\downarrow$ & LPIPS $\downarrow$ & $\eta_{\text{cfg}}$ & FVD $\downarrow$ & LPIPS $\downarrow$ \\
\midrule
0.00& 24.8 & 0.199 & 0.0 & 24.2 & 0.254 \\
0.05          & \textbf{23.8} & 0.195& 1.0       & 23.1 & 0.223 \\
0.10          & 24.5 & 0.195  & \textbf{1.5} & \textbf{22.3} & \textbf{0.219} \\
0.20          & 25.1 & \textbf{0.191}      & 2.0 & 22.7 & 0.221 \\
0.50          & 27.6 & 0.199  & \multicolumn{3}{c}{($\eta_{\mathrm{t}}{=}0.05$)} \\
\bottomrule
\end{tabular}
\end{table}

\section{Additional Experimental Setup}
\label{sec:app-setup}

\subsection{Datasets}
\label{sec:app-datasets}

\paragraph{TECO-Minecraft\citep{yan2023teco}.}
We adopt the long-context, action-conditioned prediction setup popularized by TECO~\cite{yan2023teco}.
Each video contains 300 frames, with $128 \times 128$ resolution, with per-frame action annotations.
We evaluate on 256 video clips; for long-horizon quality (FVD), each clip supplies 36 ground-truth context frames followed by 264 generated frames; for frame‑wise metrics (LPIPS, SSIM, PSNR), each clip supplies 144 observed frames followed by 156 generated frames. This set-up exactly aligns with TECO.

\paragraph{UCF-101\citep{soomro2012ucf101}.}
We use the UCF-101 dataset to demonstrate the models' capability in real-world motion.
This dataset comprises $\sim\!$13K unconstrained, unconditional action videos.
Following MCVD~\cite{voleti2022mcvd}/ExtDM~\cite{zhang2024extdm} and FAR~\cite{gu2025far}, we randomly sample 256 videos and, for each, draw 100 stochastic trajectories. 
Pixel metrics (LPIPS/SSIM/PSNR) are computed best-of-100 per video, and FVD is averaged over all 100 trajectories.

\paragraph{RealEstate10K\citep{zhou2018stereo}.}
We also perform unconditional generation experiments on an auxiliary benchmark, RealEstate10K, a dataset consisting of real-world indoor scenes. While this dataset is predominantly used in 3D tasks, we use it because it is a relatively small real-world dataset, where pretraining is feasible for our experiments. 
We use a resolution of $256 \times 256$. Since our model is orthogonal to camera-pose conditioning techniques, we simply perform unconditional prediction tasks with 16 context frames and 48 generated frames. Results are shown in Tab~\ref{tab:re10k-uncond}.

\begin{table}[htbp]
\centering
\footnotesize
\setlength{\tabcolsep}{4pt}
\renewcommand{\arraystretch}{1.1}
\caption{\textbf{Unconditional generation on RealEstate10K (256$^2$, $16{\rightarrow}48$).} 
Numbers are measured at 400k training steps with 50 denoising steps.}
\label{tab:re10k-uncond}
\begin{tabular*}{\linewidth}{@{\extracolsep{\fill}}l c c c c c@{}}
\toprule
\multirow{2}{*}{Model} & \multirow{2}{*}{Sec/F} & \multicolumn{4}{c}{\textbf{16$\rightarrow$48}} \\
\cmidrule(lr){3-6}
 &  & LPIPS$\downarrow$ & SSIM$\uparrow$ & PSNR$\uparrow$ & FVD$\downarrow$ \\
\midrule
Causal DiT-B & 1.07        & 0.172 & 0.594 & 19.35 & 101.64 \\
\midrule
\textbf{\mymodelab-B}      & 0.44        & 0.142 & 0.616 & 19.67 & 102.83 \\
\textbf{\mymodelab-B$^{E}$} & 0.45        & \underline{0.139} & \underline{0.622} & \underline{19.95} & \underline{101.61} \\
\textbf{\mymodelab-B$^{D}$} & 1.03        & \textbf{0.135} & \textbf{0.623} & \textbf{20.01} & \textbf{85.12} \\
\bottomrule
\end{tabular*}
\vspace{-0.75em}
\end{table}

\paragraph{Tokenizer.}
Following the common practice, we compress video frames into video latent, and apply diffusion models in the corresponding latent space. To compress a video, we adopt a series of VAE and DCAE models~\cite{chen2025deepcompressionautoencoderefficient}.
For Minecraft and UCF, we use the DCAE trained in FAR~\cite{gu2025far} 
; for RealEstate10K, we adopt the E2E-VAE tokenizer from \cite{leng2025repae} finetuned from VA-VAE~\cite{yao2025reconstructionvsgenerationtaming}.
\section{Model and Training Details}
\label{sec:app-impl}

\subsection{Design Details of Separable Causal Diffusion}
\label{sec:app-notation}
Let $\Enc$ denote the causal reasoning encoder and $\Dec$ the frame‑wise diffusion decoder. The encoder runs once per video frame outside the denoising loop to produce a latent $\ct$, summarizing the temporal context. Then, the decoder denoises each frame with multiple diffusion steps, conditioned on $\ct$:
\[
\ct = \Enc(x_{<t}, a_{\le t}) \quad,\qquad
\hat v^{(s)}_t = \Dec\!\bigl(x^{(s)}_t,\, s,\, \ct\bigr).
\]
Here $x^{(s)}_t$ are noisy frame latents at step $s$, and $\hat v^{(s)}_t$ is the predicted velocity/score used by the sampler. 
$\Enc$ uses frame‑wise causal attention and KV caches; $\Dec$ performs intra-frame bidirectional attention. The amortized per-frame cost is therefore
\(
O(\Enc)\;+\;\Sden\cdot O(\Dec)
\).

In our experiments, we choose $O(\Enc)\!\gg\!O(\Dec)$ for two reasons: 1) Empirically, we observe that a large portion of layer features are shareable across the denoising process (Fig.\ref{fig:layer-sim-combined}). 2) The encoder's cost is amortized across multiple denoising steps, so it can be made larger without significantly sacrificing efficiency.

\subsection{Architectures and Model Variants}
\label{sec:app-arch}
We follow the Diffusion Transformer (DiT) structure~\cite{peebles2023dit} to implement the SCD neural network. 
We follow DiT's width/head configuration for hidden size and MLP. To compare with FAR~\cite{gu2025far}, the SOTA model on Minecraft, we also adopt its FAR-M parametrization. Table~\ref{tab:model-variants-decoupled} enumerates the variants used in our experiments and reports BP/frame under $\Sden{=}50$. 

\begin{table}[htbp]
\centering
\footnotesize
\caption{\textbf{Model variants and depth split.} Depth is split into $\lenc$ causal blocks and $\mdec$ diffusion blocks. BP/frame $=\lenc + \Sden\!\cdot\!\mdec$ with $\Sden{=}50$.}
\label{tab:model-variants-decoupled}
\begin{tabular}{lccccc}
\toprule
Model & \#Blocks & Hidden & \#Heads & Params & BP / frame \\
\midrule
\textbf{DiT-B}      & $12$    & 768  & 12 & 131M & 600 \\
\midrule
\textbf{\mymodelab-B}      & $8{+}4$    & 768  & 12 & 132M & 208 \\
\textbf{\mymodelab-B$^E$}  & $12{+}4$   & 768  & 12 & 174M & 212 \\
\textbf{\mymodelab-B$^D$}  & $8{+}12$   & 768  & 12 & 217M & 608 \\
\midrule
\textbf{FAR-M}      & $12$    & 1024  & 16 & 230M & 600 \\
\midrule
\textbf{\mymodelab-M}      & $8{+}4$    & 1024 & 16 & 230M & 208 \\
\textbf{\mymodelab-M$^E$}  & $12{+}4$   & 1024 & 16 & 306M & 212 \\
\textbf{\mymodelab-M$^D$}  & $8{+}12$   & 1024 & 16 & 383M & 608 \\
\bottomrule
\end{tabular}
\end{table}

\subsection{Algorithmic Pipeline}
\label{sec:app-algorithm}
We summarize the end-to-end training and inference procedures of Separable Causal Diffusion (SCD) in Algorithms~\ref{alg:csd-train-simple} and~\ref{alg:csd-infer-simple}. 
\begin{algorithm}[t]
\caption{SCD Training}
\label{alg:csd-train-simple}
\begin{algorithmic}[1]
\Require Videos $\mathbf{x}_{1:N}$ with controls $\mathbf{a}_{1:N}$, where $N$ is the number of frames;
         temporal reasoning module $\mathcal{E}_\phi$; frame diffusion module $\mathcal{D}_\theta$;
         diffusion loss $\mathcal{L}$; noisy multi-batch size $K$.
\Repeat
  \State Choose target frame $i \in \{1,\dots,N\}$
  \State $c_i \gets \mathcal{E}_\phi\!\left(\mathbf{x}_{<i}, \mathbf{a}_{\le i}\right)$ 
  \State $\mathcal{L}_\text{step} \gets 0$
  \For{$k = 1$ to $K$}
    \State Sample $t \sim \mathcal{U}[0,1]$, $\epsilon \sim \mathcal{N}(0,I)$
    \State $x_i^{t} \gets (1-t)\,x_i + t\,\epsilon$
    \State $\hat{u} \gets \mathcal{D}_\theta\!\left(x_i^{t}, t, c_i\right)$
    \State $\mathcal{L}_\text{step} \gets \mathcal{L}_\text{step} + \mathcal{L}\!\left(\hat{u},\,x_i,\,\epsilon,\,t\right)$
  \EndFor
  \State Take a gradient step on $\nabla_{\theta,\phi}\,\bigl(\mathcal{L}_\text{step} / K\bigr)$
\Until{converged}
\end{algorithmic}
\end{algorithm}

\begin{algorithm}[t]
\caption{SCD Generation by Roll-out over Frames}
\label{alg:csd-infer-simple}
\begin{algorithmic}[1]
\Require Controls $\mathbf{a}_{1:N}$, where $N$ is the number of frames to be generated; temporal reasoning module $\mathcal{E}_\phi$; frame diffusion module $\mathcal{D}_\theta$;
         sampler with $T$ denoising steps and schedule $\{t_1,\dots,t_T\}$.
\State $\hat{\mathbf{x}} \gets [\,]$ \hfill{\small\% generated frames buffer}
\For{$i = 1,\dots,N$}
  \State $c_i \gets \mathcal{E}_\phi\!\left(\hat{\mathbf{x}}_{<i}, \mathbf{a}_{\le i}\right)$ \hfill{\small\% AR context: previously generated frames}
  \State Initialize $z^{T} \sim \mathcal{N}(0,I)$
  \For{$t = T, T\!-\!1, \dots, 1$}
    \State $\hat{u} \gets \mathcal{D}_\theta\!\left(z^{t}, t, c_i\right)$
    \State $z^{t-1} \gets \textsc{Sampler}\!\left(z^{t}, \hat{u}, t\right)$
  \EndFor
  \State $\hat{x}_i \gets z^{0}$; \quad append $\hat{x}_i$ to $\hat{\mathbf{x}}$
\EndFor
\State \Return $\hat{\mathbf{x}}_{1:N}$
\end{algorithmic}
\end{algorithm}

\section{SCD Fine-tuning Details}
\label{sec:app-finetune}

This section describes fine-tuning details omitted from the main text. We first specify the teacher, data, and architecture adaptations used to convert a pretrained bidirectional video diffusion model into our Separable Causal Diffusion (SCD), which contains a causal encoder + a frame‑wise diffusion decoder. We then detail the self-forcing rollout/distillation protocol used for post-training, and finally discuss capacity splits between encoder and decoder that highlight the flexibility of SCD.



\paragraph{Bidirectional Teacher.}
Unless otherwise noted, we fine-tune from a high-quality, bidirectional T2V checkpoint of WAN 2.1 T2V-1.3B \cite{wan2025}, whose weights are transplanted into our decoupled backbone (\S\ref{sec:app-finetune:adapt}). All training lies in the latent spaces derived from the original VAE of WAN 2.1.


\paragraph{Datasets.}
We use the text prompts from a 1M subset of VidProM~\cite{wang2024vidprom} following the same filtering process in \cite{huang2025selfforcing}. For fine-tuning with diffusion loss with our architecture, 
we use 70k synthetic data generated by WAN 2.1 T2V-14B with the above text prompts. For self-rollout training, we use the full 1M text prompts as conditions.

\paragraph{Training Specifications.}
We jointly train the encoder and decoder with the conditional flow‑matching objective (Eq.~(1) in the main text). For time step distribution, following WAN, we employ the timestep shifting $t^\prime(k, t) = \frac{kt}{1  +(k-1)t}$ and the forward interpolation is given as $x_t = ({1 - t^\prime}) x + {t^\prime} \epsilon$, $\epsilon \sim \mathcal N(0, I), t \in \mathcal U(0, 1)$. We use the AdamW~\cite{loshchilov2018decoupled} optimizer for all experiments. Detailed hyperparameters can be found in the Table~\ref{tab:finetune-hparams} and Table~\ref{tab:sf-hparams}.

\subsection{Architecture Adaptation for Decoupling}
\label{sec:app-finetune:adapt}

As described in the main paper, our decoupled backbone implements once‑per‑frame temporal reasoning in a causal encoder $\,\mathcal{E}_\phi\,$ and iterative rendering in a light frame‑wise diffusion decoder $\,\mathcal{D}_\theta$, which differs from the teacher architecture. Therefore, to align the two architectures, in practice we make two adaptations when initializing from a bidirectional teacher:

\paragraph{(i) Input reparameterization for the encoder.}
Pretrained video diffusers consume a \emph{noisy} current frame at each denoising step, while our encoder must operate on last generated frame instead of current frame. During fine‑tuning, we therefore feed the encoder a corrupted current frame $x^{\,t}_i$ at relatively \emph{high} noise levels (e.g., top 20\% of the diffusion/flow schedule), and at inference we replace it with pure Gaussian noise. This aligns the encoder’s input distribution with the teacher while preserving the decoupled compute pattern (once per frame for $\mathcal{E}_\phi$, multi‑step for $\mathcal{D}_\theta$). \textit{}

\paragraph{(ii) Layer decomposition.}
Directly treating early layers as encoder and late layers as decoder often harms generation performance from finetuning. As discussed in the main text, the early layers play an important role in converting model input scale to an internal model scale. Guided by leave‑one‑out loss probing, we allocate the first 25 layers to $\mathcal{E}_\phi$ and build $\mathcal{D}_\theta$ by combining the first 5 and last 5 layers. 

\subsection{Hyperparameters}
\label{sec:app-finetune:hparams}

\begin{table}[h]
\centering
\small
\setlength{\tabcolsep}{6pt}
\caption{\textbf{Fine‑tuning hyperparameters.} }
\label{tab:finetune-hparams}
\begin{tabular}{l l}
\toprule
Resolution / Frames & 832$\times$480 / $81$ frames\\
Batch size & 64 \\
LR / WD / Optimizer & $2\!\times\!10^{-5}$ / $0.01$ / AdamW($0.9,0.99$) \\
EMA decay & 0.99 \\
Time sampler & $\frac{5t}{1 + 4t}, t \sim \mathcal U(0, 1)$ \\
\bottomrule
\end{tabular}
\end{table}

\begin{table}[h]
\centering
\small
\setlength{\tabcolsep}{6pt}
\caption{\textbf{Self-Forcing rollout training hyperparameters.} }
\label{tab:sf-hparams}
\resizebox{\columnwidth}{!}{
\begin{tabular}{l l}
\toprule
Resolution / Frames & 832$\times$480 / $81$ frames\\
Teacher & WAN 2.1 14B\\
Teacher CFG & 3.0\\
Critic initialization & WAN 2.1 1.3B \\
Batch size & 64 \\
Student LR / WD / Optimizer & $2\!\times\!10^{-6}$ / $0.01$ / AdamW($0.0,0.99$) \\
Critic LR / WD / Optimizer & $4\!\times\!10^{-7}$ / $0.01$ / AdamW($0.0,0.99$) \\
Student EMA decay & 0.99 \\
Critic/student update ratio & 5 \\
Time sampler & $\frac{5t}{1 + 4t}, t \sim \mathcal U(0, 1)$ \\
\bottomrule
\end{tabular}
}
\end{table}
\paragraph{Throughput and latency.}
We report wall‑clock throughput (FPS) and per‑frame latency with batch size 1 on 1$\times$H100 80\,GB, charging the initial frame’s extra compute. SCD fine‑tuned from a strong T2V teacher achieves $\sim$11.1 FPS with 0.29\,s latency at 832$\times$480 while retaining competitive VBench scores; the self‑forcing baseline at the same scale reaches 8.9 FPS and 0.45\,s latency.(Table~\ref{tab:vbench-finetune}). 

\begin{table}[h]
\centering
\footnotesize
\setlength{\tabcolsep}{6pt}
\caption{\textbf{Frame-Autoregressive Text‑to‑Video on VBench (832$\times$480, batch 1)}. Reported throughput includes first‑frame overhead.}
\label{tab:vbench-finetune}
\begin{tabular}{lcccc}
\toprule
Model & FPS $\uparrow$ & Latency (s) $\downarrow$ & Total $\uparrow$ & Quality / Semantic $\uparrow$ \\
\midrule
Self Forcing & 8.9 & 0.45 & 84.26 & 85.25 / 80.30 \\
\textbf{SCD (Ours)} & \textbf{11.1} & \textbf{0.29} & 84.03 & 85.14 / 79.60 \\
\bottomrule
\end{tabular}
\end{table}

\subsection{Flexibility of Separable Causal Diffusion}
\label{sec:app-finetune:flex}

A practical benefit of SCD is that temporal reasoning capacity and per‑frame rendering capacity can be traded \emph{independently}. Let total depth be $\lenc{+}\mdec$ with encoder depth $\lenc$ (causal reasoning, amortized once per frame) and decoder depth $\mdec$ (frame‑wise denoising, repeated $S$ steps in inference, where $S=4$ in standard self-forcing settings). For fixed parameters:

\begin{itemize}[leftmargin=1.25em,itemsep=2pt,topsep=2pt]
  \item \textbf{Encoder‑heavier variants} slightly increase reasoning cost per frame but systematically improve motion/layout adherence and long‑horizon stability; it only slightly reduce the throughput since the encoder runs once per frame.
  \item \textbf{Decoder‑heavier variants} improve per‑frame detail at the cost of $S{\times}\mdec$ block passes per frame; quality gains can be notable when targeting high‑fidelity or very few denoising steps, but latency rises accordingly.
\end{itemize}

\end{document}